\documentclass[conference]{IEEEtran}
\usepackage{times}

\usepackage[sort&compress,numbers]{natbib}
\usepackage{multicol}
\usepackage[bookmarks=true]{hyperref}

\usepackage{wrapfig}
\usepackage{url}            
\usepackage{amsfonts}       
\usepackage{nicefrac}       
\usepackage{microtype}      
\usepackage{xcolor}         
\usepackage{amsmath,amssymb} 
\usepackage{pifont}  
\usepackage{makecell}
\usepackage{diagbox}
\usepackage{booktabs}
\usepackage{graphics} 
\usepackage{adjustbox}
\usepackage{colortbl}
\usepackage{xcolor}
\usepackage{empheq}
\usepackage{bm}
\usepackage{bbding} 
\usepackage[textwidth=1in]{todonotes}
\usepackage{diagbox}
\usepackage[small]{caption}

\newcommand{\xmark}{\ding{55}}%
\definecolor{lightblue}{HTML}{DBE9FC}
\definecolor{lighterblue}{HTML}{f1fcfe}
\definecolor{darkblue}{HTML}{6c8ebf}

\usepackage[capitalize]{cleveref}
\usepackage{float}
\usepackage{booktabs}
\usepackage{multirow}
\usepackage{mathrsfs}
\usepackage[utf8]{inputenc}
\usepackage{pifont}
\usepackage{threeparttable}
\usepackage[lined,boxed,commentsnumbered,ruled,vlined]{algorithm2e}
\usepackage{tcolorbox}
\usepackage[frozencache,cachedir=.]{minted}
\SetKwComment{Comment}{$\triangleright$\ }{}

\hypersetup{
    colorlinks,
    linkcolor={red!50!black},
    citecolor={blue!50!black},
    urlcolor={blue!80!black}
}

\let\subparagraph\paragraph
\usepackage[compact]{titlesec}
\usepackage{amsthm}

\theoremstyle{definition}
\newtheorem{assumption}{Assumption}

\newtheorem{definition}{Definition}

\DeclarePairedDelimiterX{\infdivx}[2]{(}{)}{%
  #1\;\delimsize\|\;#2%
}

\newcounter{RNum}

\newcommand{\fref}[1]{Figure~\ref{#1}}
\newcommand{\sref}[1]{Section~\ref{#1}}
\newcommand{\tref}[1]{Table~\ref{#1}}
\newcommand{\appref}[1]{Appendix~\ref{#1}}
\newcommand{\defref}[1]{Definition~\ref{#1}}

\newcommand{\myparagraph}[1]{\noindent\textbf{#1}~}

\makeatletter
\apptocmd\@maketitle{{\myfigure{}\par}}{}{}
\newcommand{\removelatexerror}{\let\@latex@error\@gobble}
\makeatother
\newcommand\blfootnote[1]{%
  \begingroup
  \renewcommand\thefootnote{}\footnote{#1}%
  \addtocounter{footnote}{-1}%
  \endgroup
}

\def\model{IVNTR}

\begin{document}

\newcommand\myfigure{%
\centering
\noindent
\includegraphics[width=0.95\textwidth]{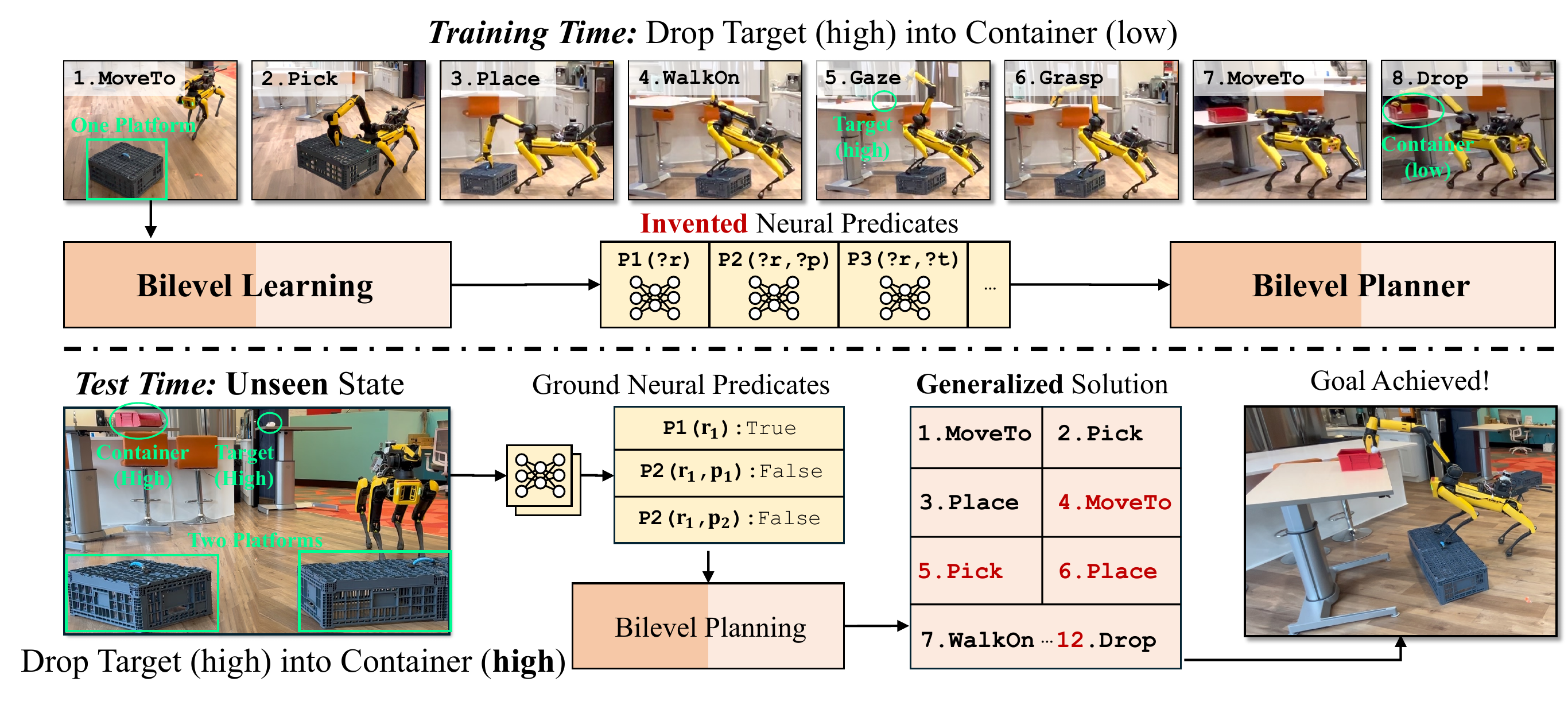}
\captionof{figure}{(Top) Our bilevel learning framework invents \textbf{neural} predicates from training demonstrations (with one platform), which enable the learning of a hybrid bilevel planner.
(Bottom) The invented predicates realize \textbf{zero-shot} compositional generalization over objects (with two platforms), where a longer solution with different action compositions is generated.
Continuous action parameters are omitted for simplicity.}
\label{fig:teaser}
\setcounter{figure}{1}
}

\title{Bilevel Learning for Bilevel Planning}

\author{\authorblockN{Bowen Li\authorrefmark{1}\authorrefmark{2},
Tom Silver\authorrefmark{3},
Sebastian Scherer\authorrefmark{1}, and
Alexander Gray\authorrefmark{2}}
\authorblockA{\authorrefmark{1}Carnegie Mellon University, \authorrefmark{2}Centaur AI Institute, \authorrefmark{3}Princeton University}}

\maketitle

\begin{abstract}
A robot that learns from demonstrations should not just imitate what it sees---it should understand the high-level concepts that are being demonstrated and generalize them to new tasks.
Bilevel planning is a hierarchical model-based approach where predicates (relational state abstractions) can be leveraged to achieve compositional generalization.
However, previous bilevel planning approaches depend on predicates that are either hand-engineered or restricted to very simple forms, limiting their scalability to sophisticated, high-dimensional state spaces.
To address this limitation, we present \model{}, the first bilevel planning approach capable of learning neural predicates directly from demonstrations.
Our key innovation is a neuro-symbolic bilevel learning framework that mirrors the structure of bilevel planning.
In \model{}, symbolic learning of the predicate ``effects" and neural learning of the predicate ``classifiers" alternate, with each providing guidance for the other.
We evaluate \model{} in six diverse robot planning domains, demonstrating its effectiveness in abstracting various continuous and high-dimensional states.
While most existing approaches struggle to generalize (with $<35\%$ success rate), our \model{} achieves an average success rate of $77\%$ on unseen tasks.
Additionally, we showcase \model{} on a mobile manipulator, where it learns to perform real-world mobile manipulation tasks and generalizes to unseen test scenarios that feature new objects, new states, and longer task horizons.
Our findings underscore the promise of learning and planning with abstractions as a path towards high-level generalization.
Project website: \url{https://jaraxxus-me.github.io/IVNTR/}.
\blfootnote{$^\dagger$Work was partly done during internship at Centaur AI Institute. Correspondence to \{bowenli2,basti\}@andrew.cmu.edu.}
\end{abstract}

\IEEEpeerreviewmaketitle

\section{Introduction}

Imitation learning has made significant recent strides~\cite{mandlekar2023mimicgen,chi2023diffusionpolicy,zhao2023learning,wang2024equivariant,yang2024equibot}, but generalization remains an open challenge, especially when new tasks require recomposing high-level concepts that are only implicit in the training data~\cite{mao2024planning,li2024logicity}.
In \fref{fig:teaser}, a robot has seen demonstrations of stepping onto a platform to grasp an object, and other demonstrations of dropping the object into a container.
Now, faced with a new task where the container is also elevated, the robot should first move the two platforms appropriately, step onto one of them to grasp the object, and finally step onto the other to drop the object.
Note that the platform arrangements must be completed before grasping the target, since the robot cannot move platforms with its hand full.
This kind of learning and reasoning requires compositional generalization (new objects); sequential generalization (new and longer action sequences); and long-horizon planning with continuous state and action spaces with sparse feedback (goals).
In sum, the robot should not just \emph{imitate} demonstrations, but also \emph{understand and leverage} the high-level concepts within the low-level states that are being demonstrated.

One promising direction to address these challenges is to learn and plan with \textit{abstractions}~\cite{li2006towards,abel2018state,konidaris2018symbols,wonglearning,curtis2022discovering,yang2024guidinglonghorizontaskmotion,shah2024reals}.
In this work, we continue a line of recent inquiry on learning abstractions for \emph{bilevel planning}~\cite{silver2021operator,silver2022skills,silver2023predicateinvent,chitnis2021glib,kumar2024practice,kumar2023predict,liang2024visualpredicator}.
In bilevel planning, continuous low-level states are mapped into a symbolic relational state space defined by \emph{predicates} such as \texttt{Viewable(?robot,?target)} or \texttt{On(?robot,?platform)}.
Planning proceeds jointly in the symbolic high-level space and the continuous low-level space.
The key idea is that this hybrid planning can be more efficient and effective than reasoning solely in the low-level space.

The performance of bilevel planning depends substantially on the predicates used to define the abstract state space~\cite{silver2023predicateinvent}.
To avoid the need for a human engineer to manually define predicates for every new domain, recent work has considered \emph{learning predicates} from data~\cite{kulick2013active,konidaris2018skills,silver2023predicateinvent,li2023embodied,han2024interpret,liang2024visualpredicator}.
Broadly, three approaches have emerged.
The most direct one relies on human feedback (labels or guidance) during the predicate learning process~\cite{li2023embodied,migimatsu2022grounding,han2024interpret}, which is labor-intensive and does not guarantee useful abstractions for planning~\cite{silver2023predicateinvent}.
The second approach \textit{invents} predicates with surrogate objectives that are easy to optimize, \textit{e.g.}, reconstruction loss~\cite{asai2018latplan_prop,asai2019latplan_fol,asai2021latplanpddl} or bisimulation~\cite{curtis2022discovering,hansen2022bisimulation}.
While these methods simplify learning, they complicate planning due to the mismatch between the surrogate objectives and the actual planning goals~\cite{silver2023predicateinvent}.
The third approach directly invents predicates for efficient planning, making planning ``easy" but learning ``hard," as objectives like \emph{total-planning-time} and \emph{expected-planning-success} are difficult to optimize~\cite{silver2023predicateinvent}.
To address this, previous works have used program synthesis with classical grammars~\cite{silver2023predicateinvent} and foundation model-based techniques~\cite{liang2024visualpredicator,athalye2024predicate}.
However, in both cases, the predicates are invented from programmatic and pre-defined classifiers, which are limited in flexibility and scalability.

Our main contribution is b\textbf{I}le\textbf{V}el lear\textbf{N}ing from \textbf{TR}ansitions (\model{}), the first approach capable of learning \emph{neural} predicates that are optimized for efficient and effective bilevel planning.
Since directly incorporating the planning objective into network training is challenging, our \model{} instead constructs a candidate neural predicate pool, which is later subselected~\cite{silver2023predicateinvent}.
The key insight behind our approach is to center learning around the \emph{effects} of predicates, which provide two major benefits: 
(1) they enable the derivation of supervision labels for \textit{transition} pairs, yielding a well-structured learning objective for training the neural network; 
and (2) the inherent sparsity of predicate effects, combined with neural learning signals, facilitates efficient symbolic learning of their structure. 
To this end, \model{} presents a novel bilevel learning framework, inspired by the structure of bilevel planning itself.
Similar to the alternation between high-level symbolic search and low-level neural sampling in bilevel planning, \model{} interleaves symbolic effect learning and neural classifier learning in an iterative process.
In each iteration, the symbolic learning proposes a candidate predicate effect across different actions, which provides labels for neural learning on transition pairs. 
Once the neural classifier converges, its validation loss guides the symbolic learning to propose the next candidate that could minimize the loss in the new iteration.
This iterative bilevel learning ultimately yields a compact set of neural predicates, which are then selected to optimize the planning objective~\cite{silver2023predicateinvent}.
The final set of invented predicates seamlessly integrates into operator and sampler learning frameworks~\cite{chitnis2021nsrt,silver2023predicateinvent}, ultimately forming a fully functional bilevel planner.

To evaluate the effectiveness of \model{}, we conduct extensive experiments across six diverse robot planning domains. 
These domains feature a wide range of low-level state representations, from SE(2) and SE(3) poses to high-dimensional point clouds. 
Furthermore, as shown in \fref{fig:teaser}, by leveraging relational predicates and AI planning, \model{} zero-shot generalizes to tasks with unseen entity compositions.
Finally, we deploy \model{} on a quadruped mobile manipulator (Boston Dynamics Spot) for two long-horizon mobile manipulation tasks. 
The learned predicates successfully abstract complex continuous states into representations compatible with the AI planner, while also providing actionable guidance for the samplers. 
We believe \model{} represents a pivotal step towards learning high-level abstractions from sophisticated low-level states.

\section{Problem Formulation}\label{sec:problem}

We propose a method that uses an offline demonstration dataset to learn planning \textit{abstractions} that generalize to test tasks with unseen objects and action compositions.
In this section, we describe the formal problem setting.
We follow the notation system introduced in previous work~\cite{silver2023predicateinvent}; see \appref{app:notation} for a complete notation glossary.

Planning problems are defined within a certain \textit{planning domain} $\langle \Lambda, \mathcal{C}, f, \Psi_{\mathrm{g}}, \Psi_\mathrm{sta} \rangle$ with a task distribution $\mathcal{T}$, where we can sample a \textit{planning task} $T\sim\mathcal{T}=\langle \mathcal{O}, \mathbf{x}_0, g\rangle$.

\begin{figure}[!t]
	\centering
	\includegraphics[width=1\columnwidth]{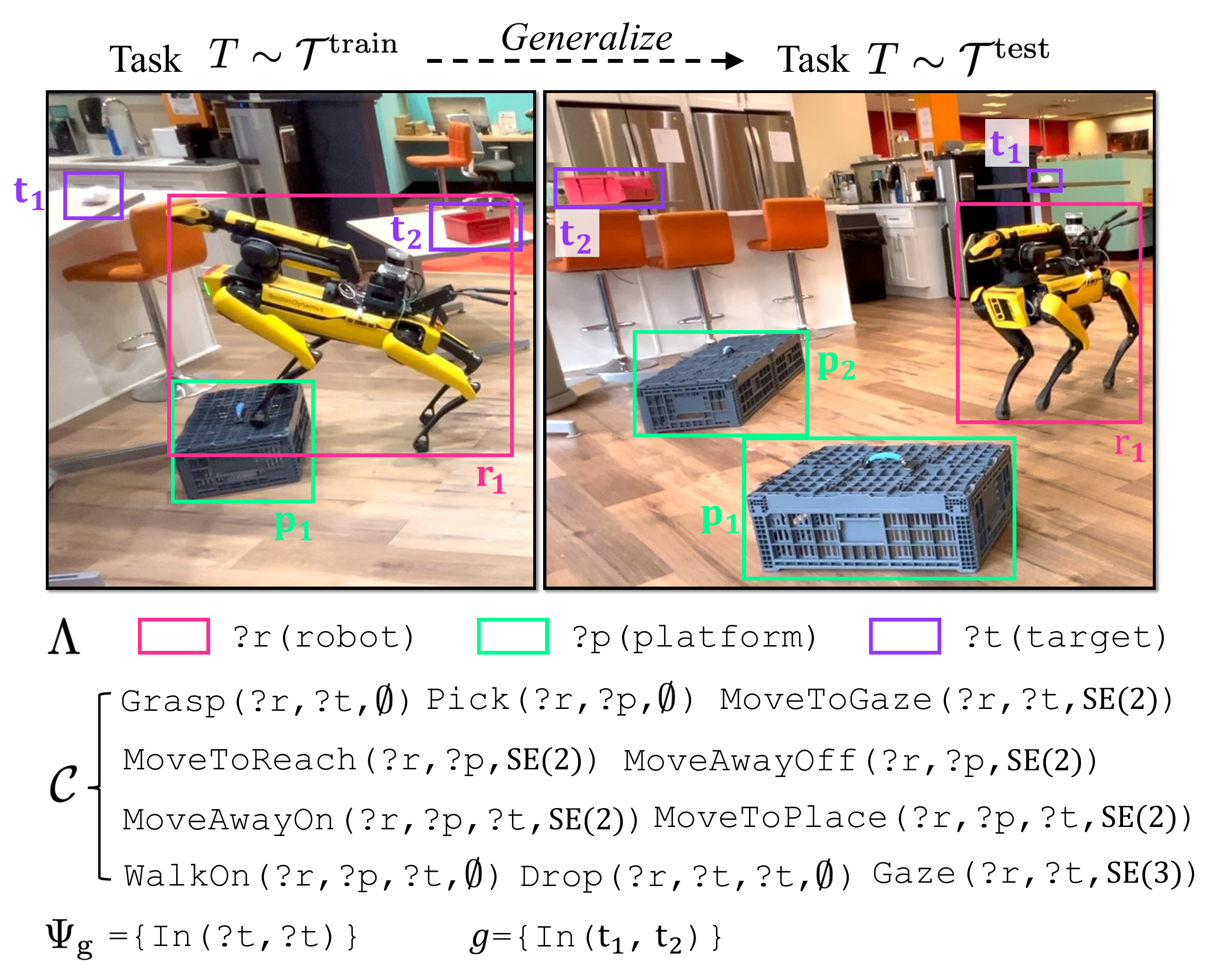}
	\caption{The Climb-Transport domain is presented as a running example. We have displayed one typical training and one test task on the top. The types, actions, and provided predicates are shown at the bottom.}
	\label{fig:running_example}
\end{figure}

$\Lambda$ is a finite set of object \textit{types} $\lambda\in\Lambda$.
For example, the Climb-Transport domain depicted in \fref{fig:running_example} has three object types: $\Lambda=\{\mathrm{robot} (\mathtt{r}), \mathrm{platform}(\mathtt{p}), \mathrm{target}(\mathtt{t})\}$.
Each type is associated with a set of \emph{features} that characterize the state of an object of that type.\footnote{Unlike previous work~\cite{silver2023predicateinvent}, we do not assume that features are scalars; high-dimensional images and point clouds are also allowed.}
For example, $\mathrm{robot}$ has features ``BasePose'', ``HandPose'', and ``GripperOpenPercent'', among others.
A specific \textit{task} $T$ is characterized by objects $\mathcal{O}=\{\mathtt{o}_1,\mathtt{o}_2,\cdots,\mathtt{o}_N\}$, each associated with one type in $\Lambda$.
Objects are fixed within tasks but vary between tasks.
The state of a task $\mathbf{x}\in\mathcal{X}$ is defined by an assignment of feature values to all objects in the task.
For simplicity of exposition, we assume that a state with $N$ objects can be represented as a matrix $\mathbf{x} \in \mathbb{R}^{N\times K}$ for some domain-specific constant $K$; however, we show in experiments that our approach can be applied to more sophisticated object-centric state representations as well.

The action space for a domain is characterized by a set of $M$ \textit{parametrized controllers} $\mathcal{C} = \{\mathtt{C}_1, \mathtt{C}_2, \cdots, \mathtt{C}_M\}$, each of which has an object type signature $(\lambda_1, \lambda_2, \cdots, \lambda_{v})$ and a continuous parameter space $\Omega$.
For example, in \fref{fig:running_example}, \texttt{MoveToReach} has type signature $(\mathrm{robot}, \mathrm{platform})$, and continuous parameters $\Omega = \text{SE}(2)$ defining an offset 2D pose for the robot relative to the platform.
A \emph{ground action} is a controller with fully specified parameters, e.g., $\texttt{MoveToReach}(\mathtt{r}_1, \mathtt{p}_1, \omega)$ for a certain $\omega \in \Omega$.
We use underline notation to represent grounding: $\underline{\mathtt{C}}$ is a certain ground action.
A \emph{lifted action} is controller with object parameter placeholders, which are typically prefixed with ?, e.g., $\texttt{MoveToReach}(\mathrm{?r}, \mathrm{?p}, \cdot)$.
States and actions are related through
a known transition function $f(\mathbf{x}, \underline{\mathtt{C}}) \mapsto \mathbf{x}'$, which the robot can use to plan.

A \emph{predicate} $\psi$ is defined by an object type signature $(\lambda_1, \lambda_2, \ldots, \lambda_{u})$ and 
a classifier $\theta_{\psi}: \mathcal{X}\times\mathcal{O} \to \{\mathrm{True}, \mathrm{False}\}$, where $\theta_{\psi}(\mathbf{x}, (\mathtt{o}_1, \ldots, \mathtt{o}_{u}))$ evaluates the truth value of a ground predicate based on the continuous features of the input objects.
For example, the predicate \texttt{In} has type signature $(\mathrm{target}, \mathrm{target})$ and a classifier that uses the poses and shapes of two targets to determine whether one is ``in'' the other.
A \emph{ground predicate} $\underline{\psi}$ has fully specified objects.
For simplicity, we denote
$\theta_{\underline{\psi}}(\mathbf{x}) \triangleq \theta_{\psi}\left(\mathbf{x}, \left(\mathtt{o}_1, \ldots, \mathtt{o}_{u}\right)\right).$
A \emph{lifted predicate} has placeholders for objects, e.g., $\texttt{In}(\texttt{?t}, \texttt{?t})$.

Following previous work~\cite{silver2023predicateinvent}, we assume that a small set of \emph{goal predicates} $\Psi_G$ is known and used to characterize task goals.
In particular, a goal $g$ is defined by a set of ground predicates that must evaluate to True in a state for the goal to be satisfied.
For example, the goal in Figure~\ref{fig:running_example} has only one ground predicate, $\texttt{In}(\texttt{t}_1,\texttt{t}_2)$.
In this work, we make an additional assumption that any relevant \emph{static predicates} $\Psi_\mathrm{sta}$ are known.
A predicate is static if its evaluation never changes within a task (see \appref{app:domain_details} for examples).
Conversely, a predicate is \textit{dynamic} if its evaluation could change within a task; examples are provided later in \defref{def:op}.

A solution to a task is a plan $\pi=[\underline{\mathtt{C}}_1, \underline{\mathtt{C}}_2, \cdots, \underline{\mathtt{C}}_H]$, that is, a sequence of $H$ ground actions such that successive application of the transition model $\mathbf{x}_i=f\left(\mathbf{x}_{i-1}, \underline{\mathtt{C}}_i\right)$ on each $\underline{\mathtt{C}}_i \in \pi$, starting from $\mathbf{x}_0$, results in a final state $\mathbf{x}_H$ where $g$ holds.
For instance, the plan depicted in \fref{fig:teaser} bottom right finally leads to the state where $\texttt{In}(\texttt{t}_1,\texttt{t}_2)$ holds.
In sum, to generate the plan $\pi$ for a task, in each state, the robot needs to predict:
(1) the action class $\mathtt{C}\in\mathcal{C}$,
(2) the objects as the discrete parameters,
and 
(3) the continuous parameter $\omega\in\Omega$.

During training, the robot has access to an offline demonstration dataset $\mathcal{D} =\{({T}_i, \pi_i)\}_{i=1}^B$, which consists of $B$ task and solution pairs sampled from the task distribution $\mathcal{T}^\mathrm{train}$.
Note that since the transition function $f$ is known and deterministic, we can also recover the intermediate states from $\mathbf{x}_0$.
During test time, the robot is required to solve held-out tasks sampled from a different \textit{test} distribution $\mathcal{T}^\mathrm{test}$.
In practice, for the sake of evaluating generalization, test tasks typically contain new and more objects than training tasks.
For example, as shown in \fref{fig:running_example}, all training tasks only have $1$ platform, but during test, there are $2$ platforms.
The difference in object compositions could result in different lengths of plans with a different action order, requiring the method to \textit{generalize} by understanding the implicit concepts present in the training demonstrations.

\section{Bilevel Planning}\label{sec:bilevel_planning}

\begin{figure*}[!t]
	\centering
	\includegraphics[width=2\columnwidth]{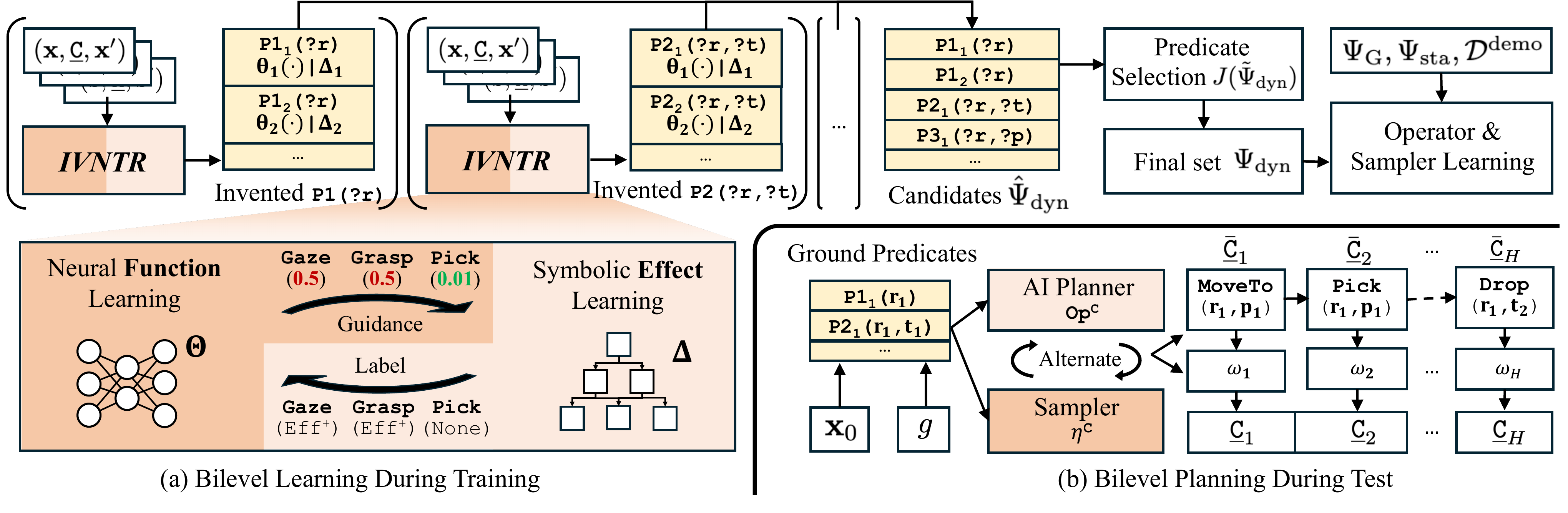}
	\caption{(a) Overview of \model{} during training. Given transition pairs in the continuous space, \model{} invents neural predicates with different arguments parallelly, resulting in a candidate set. A subset that minimizes the planning objective $J(\cdot)$ is selected from the candidates, which serves as the final $\Psi_{\mathrm{dyn}}$. With the complete predicate set, sampler and operator learning can be achieved.
    (b) Bilevel planning with operators and samplers during test. Compositional ground predicates serve as inputs to the AI planner and provide guidance to the samplers.
    }
	\label{fig:overview}
\end{figure*}

In this work, we propose a method for learning predicates that can be used for \emph{bilevel planning}.
We now provide a brief review of bilevel planning and refer to other references for a more in-depth discussion~\cite{chitnis2021nsrt,liang2024visualpredicator,silver2023predicateinvent,silver2022skills,li2023embodied,kumar2023predict,kumar2024practice,silver2021operator,garrett2021integrated}.

Bilevel planning uses relational abstractions to achieve sequential and compositional generalization.
The two principal abstractions are \emph{predicates} (state abstractions) and \emph{operators} (action abstractions).
Bilevel planning also uses relational \emph{samplers} to generate possible ground actions from operators.
The key idea is that planning jointly in the abstract transition system and the low-level transition system can be much more efficient than planning in the low-level transition space only.

\begin{definition}[Operator]\label{def:op}
    The \textit{operator} for a parametrized controller $\mathtt{C}$ is a tuple, $\mathtt{Op}^\mathtt{C} = \langle \mathtt{Var}, \mathtt{Pre}, \mathtt{Eff}^+, \mathtt{Eff}^- \rangle$, where
    $\mathtt{Var}$ is a tuple of object placeholders matching the type signature of $\mathtt{C}$, and
  $\mathtt{Pre},\,\mathtt{Eff}^+,\,\mathtt{Eff}^- \subseteq \Psi$, respectively \emph{preconditions}, \emph{add effects}, and \emph{delete effects}, are each a set of lifted predicates defined with variables in $\mathtt{Var}$. $\Psi$ is a predicate set.
\end{definition}
For example, the operator for $\mathtt{Grasp(?r,?t)}$ could be:
\begin{align*}
& \mathtt{Pre}=\{\mathtt{HandEmpty(?r)},\mathtt{HandSees(?r,?t)}\},\\
& \mathtt{Eff}^+=\{\mathtt{Holding(?r,?t)}\},\\
& \mathtt{Eff}^-=\{\mathtt{HandEmpty(?r)},\mathtt{HandSees(?r,?t)}\}.
\end{align*}
Given a task $T = \langle \mathcal{O}, \mathbf{x}_0, g\rangle$, bilevel planning (Figure~\ref{fig:overview}b) starts by using predicates to generate an \emph{abstract state} consisting of all ground predicates with objects $\mathcal{O}$ whose classifiers evaluate to True in $\mathbf{x}_0$.
The initial ground predicates, together with the operator set and goal $g$, can then be input to an AI planner~\cite{helmert2006fast} to generate a plan \textit{skeleton}, $\bar{\pi}$ with partially grounded actions with unspecified continuous parameters, $\underline{\bar{\mathtt{C}}}$.
To refine the actions in this skeleton $\underline{\bar{\mathtt{C}}}\in\bar{\pi}$ into fully grounded $\underline{{\mathtt{C}}}$ with the continuous parameters $\omega$, bilevel planning leverages \textit{samplers}.
\begin{definition}[Sampler]
    The \textit{sampler} $\eta^\mathtt{C}$ for a planning operator $\mathtt{Op}^\mathtt{C}$ with $v$ object placeholders is a conditional distribution $\omega \sim \eta^\mathtt{C}(\cdot \mid \mathbf{x}, (\mathtt{o}_1, \ldots, \mathtt{o}_{v}))$ that proposes continuous action parameters for $\mathtt{C}((\mathtt{o}_1, \ldots, \mathtt{o}_{v}), \cdot)$ given a state $\mathbf{x}$.
\end{definition}
Note that unlike the deterministic operators, samplers are usually stochastic and may fail in certain steps. 
Thus, bilevel planning alternates between the AI planner and samplers, using the predicate classifiers $\theta_\Psi$ as ``guidance" in each step to compensate for potential sampling failures.

Assuming we have a complete predicate set, previous work has studied the problem of learning \textit{operators}~\cite{chitnis2021nsrt} and \textit{samplers}~\cite{kumar2024practice,silver2022skills} from the demonstration dataset $\mathcal{D}$.
Since the predicates, learned operators, and samplers are \textit{relational}, they can be generally applied to held-out test tasks sampled from $\mathcal{T}^\mathrm{test}$.
However, with an insufficient predicate set---for example, with only $\Psi_\mathrm{G}$ and $\Psi_\mathrm{sta}$---bilevel planning can be intractably slow~\cite{silver2023predicateinvent}.
We next introduce the \model{} framework, which closes this gap by automatically inventing dynamic predicates for efficient bilevel planning.

\section{Methodology}\label{sec:ivntr}

The problem of inventing dynamic predicates $\Psi_\mathrm{dyn}$ can be decomposed into \emph{symbolic learning}---how many predicates should be invented, and with what type signatures---and \emph{classifier learning}, determining $\theta_\psi$ for each invented predicate $\psi \in \Psi_\mathrm{dyn}$.
Previous approaches~\cite{liang2024visualpredicator,silver2023predicateinvent} address these problems via a ``define-then-select" two-stage pipeline.
In the first stage, for each predicate candidate $\hat{\psi}$ with known variable types, its classifer is pre-defined via program synthesis~\cite{silver2023predicateinvent} or pre-trained foundation models~\cite{liang2024visualpredicator}.
These candidates form a large predicate pool $\hat{\Psi}_\mathrm{dyn}$.
In the second stage, to subselect predicates from the pool, each candidate predicate set $\tilde{\Psi}_\mathrm{dyn}\subseteq\hat{\Psi}_\mathrm{dyn}$ is scored with a function $J(\tilde{\Psi}_\mathrm{dyn})$ that measures both planning \emph{efficiency} and \emph{effectiveness}.
A key limitation of this ``define-then-select'' pipeline is that the classifiers within $\hat{\Psi}_\mathrm{dyn}$ are restricted to a relatively simple set.
Scaling to more general classifiers, e.g., neural networks, is nontrivial, since $J(\tilde{\Psi}_\mathrm{dyn})$ is highly non-differentiable.
To address this, we propose \model{}, a ``learn-then-select'' approach that leverages \emph{bilevel learning}.

As depicted in \fref{fig:overview} (a), given the domain types $\Lambda$, \model{} enumerates all possible typed variable compositions (with maximum input arity).
Since the input features for the predicate classifier depend on its argument types, \model{} invents predicates with different arguments group by group parallelly.
For the invention of each group, \model{} draws inspiration from bilevel planning itself, where planning alternates between the symbolic level and the low level.
Similarly, \model{} interleaves \emph{symbolic learning} and \emph{neural learning}, with each providing guidance for the other.
Specifically, symbolic learning proposes \emph{effect vectors} that represent the add and delete effects for candidate predicates across all operators.
Neural learning uses these effect vectors to provide supervision for classifier learning.
The validation loss in neural learning feeds back into symbolic learning, and the process repeats.
In this section, we describe these steps in detail via the exemplar predicate group $\psi\in\Psi^{\mathtt{Var}}$, with the typed variables $\mathtt{Var}=\mathtt{(?r,?t)}$.

\subsection{Effect Vectors as Supervision for Neural Learning}\label{sec:neural_learning}
Suppose we had access to the symbolic components of a predicate $\psi$, but did not yet know its classifier $\theta_\psi$.
Suppose further that we had knowledge of all appearances of $\psi$ in the effect sets ($\mathtt{Eff}^+, \mathtt{Eff}^-$) for each operator $\mathtt{Op}^\mathtt{C}$.
We now describe how this knowledge---which we do not actually have, but which will be suggested by symbolic learning---can be used for supervised learning of the classifier $\theta_\psi$.

Recall that we have access to demonstrations $\mathcal{D}$, and for each demonstrated task $T$, we can recover the solution trajectory, $[\mathbf{x}_0,\underline{\mathtt{C}}_0,\mathbf{x}_1,\underline{\mathtt{C}}_1,\cdots,\underline{\mathtt{C}}_{H-1},\mathbf{x}_H]$.
If we knew the initial state ground predicates $\underline{\psi}$ in $\mathbf{x}_0$, then we could immediately recover all ground predicates for all the states in the trajectories by chaining together the operator effects.
Then, a simple binary classification framework could easily address our neural learning problem.
However, we do not have access to the initial state ground predicates---we only have access to operator effects---so we do not have direct knowledge of the abstract states in the demonstration.
Nonetheless, we can still provide supervision for neural learning by leveraging the ground predicates that are added, deleted, or stay unchanged in each \textit{transition pair}.
We provide this supervision by way of \emph{predicate effect vectors}, including the \emph{lifted effect vector} for a domain, and the \emph{ground effect vector} for a transition.

\begin{figure}[!t]
	\centering
	\includegraphics[width=1.01\columnwidth]{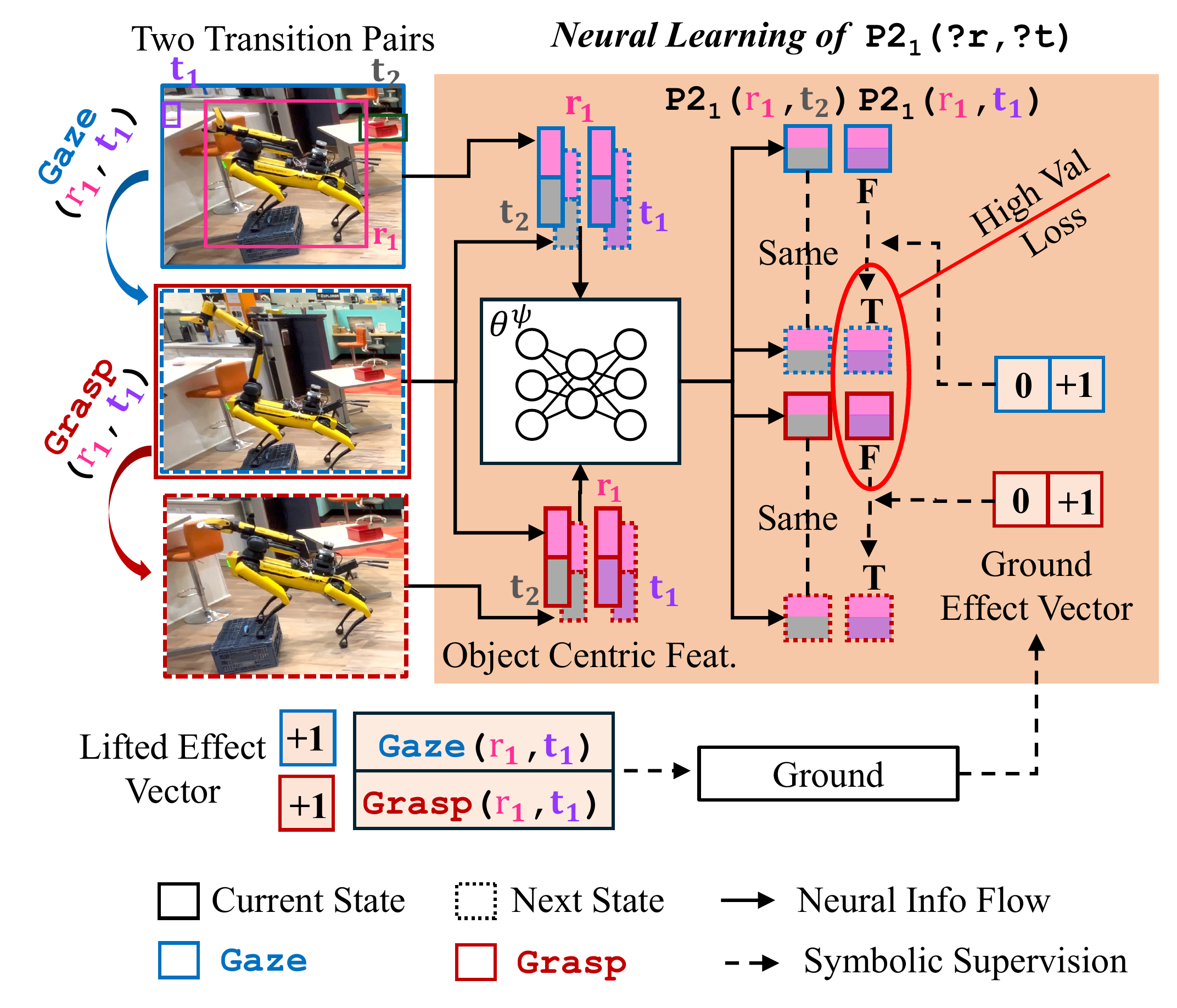}
	\caption{Detailed neural learning process for predicate $\mathtt{P2_1(?r,?t)}$.
    From the demonstration dataset, we display two transition pairs (one for each action, in total four states) on the left.
    The neural network takes object centric features as input, predicting ground predicates (in total eight values).
    At the bottom, we display an example lifted effect vector for action Grasp, Gaze as $\Delta^\psi_t=[+1,+1]$.
    With the ground effect vector, supervisions can be derived on the predicted values.
    Due to the unreasonable effect supervisions, the intermediate state is labeled as both \texttt{True} and \texttt{False}, resulting in high validation loss.
    }
	\label{fig:method_1}
\end{figure}

\begin{definition}[Lifted Effect Vector]\label{def:effect_vec}
    The \textit{lifted effect vector} for predicate $\psi$ is $\Delta^{\psi} = [\delta^{\psi}_{1}, \cdots, \delta^{\psi}_{M}] \in \{-1, 0, 1\}^M$ where:
    \[ \delta^{\psi}_{i} = \begin{cases} 
      1 & \psi \in \mathtt{Eff}^+ \text{ for } \mathtt{C}_i \\
      -1 & \psi \in \mathtt{Eff}^- \text{ for } \mathtt{C}_i \\
      0 & \text{otherwise.} 
   \end{cases}
\]
For example, in \fref{fig:method_1}, the effect vector $[+1,+1]$ specifies that predicate $\psi=\mathtt{P2_1(?r,?t)}$ is the add effect for both $\mathtt{Gaze(?r,?t)}$  and $\mathtt{Grasp(?r,?t)}$\footnote{Each predicate here can appear at most once in the effect sets, but this doesn't affect the representation capability of the final predicate set.}.
\end{definition}
The lifted effect vector is a favorable symbolic representation of a lifted predicate, since its shape doesn't depend on task object compositions and can thus be learned more efficiently, as we will see.
However, to train the neural classifier on the transition pairs, we will need to derive supervision on \textit{ground predicates}, which is achieved by the \emph{ground effect vector}.
\begin{definition}[Ground Effect Vector]\label{def:ground_effect_vec}
    Let $\mathcal{O}$ be the object set in a task $T$, $\underline{\mathtt{C}}_i$ be a ground action with objects $\mathcal{O}_{\underline{\mathtt{C}}_i}\subseteq\mathcal{O}$, then the ground effect vector $\bm{t}^{\psi, \underline{\mathtt{C}}_i} = [t_1, \cdots, t_P] \in \{-1, 0, 1\}^P$ for predicate $\psi$ grounded on $\mathcal{O}$ is defined as:
    \[ t_p
    \;=\;
    \begin{cases}
    \delta^{\psi}_i, 
    & \text{if } \mathcal{O}_{\underline{\psi}_p} \subseteq \mathcal{O}_{\underline{\mathtt{C}}_i}, \\
    0,     
    & \text{otherwise},
    \end{cases}
\]
    where $\underline{\psi}_p$ denotes the $p$-th atom with objects $\mathcal{O}_{\underline{\psi}_p}$, among the total of $P$ ground predicates. For example, in \fref{fig:method_1}, ground effects for $\mathtt{P2_1(r_1,t_1)}$ will be $+1$ for both ground actions $\mathtt{Gaze(r_1,t_1)}$ and $\mathtt{Grasp(r_1,t_1)}$, while ground effects for $\mathtt{P2_1(r_1,t_2)}$ will be $0$, as $\mathtt{(r_1,t_2)} \not\subseteq \mathtt{(r_1,t_1)}$.
\end{definition}
Importantly, this implies that the ``value" of the (potentially) non-zero entry of all ground effects from the action class $\mathtt{C}_i$ equals $\delta^{\psi}_i$, while the ``position" of the non-zero entry is decided by the object set $\mathcal{O}_{\underline{\mathtt{C}}}$ and the predicate grounding (see \appref{app:assumption} and \appref{app:same_type} for more explanations).

Now, we are able to train the neural classifier $\theta_\psi$ on the transition pairs with supervisions from $\Delta^\psi$.
Specifically, consider a transition pair: $(\mathbf{x}, \underline{\mathtt{C}}_i, \mathbf{x}')$, 
we first construct the ground effect vector $\bm{t}^{\psi,\underline{\mathtt{C}}_i}$ using $\delta^\psi_i\in\Delta^\psi$.
Then, the following supervised learning objective can be established for $\theta_\psi$:
\begin{equation}
    \label{eqn:loss}
    \begin{aligned}
    \mathcal{L}(\mathbf{x}, \mathbf{x}', \theta_\psi) 
    = \mathcal{L}_\mathrm{zero} + \mathcal{L}_\mathrm{one}
    \end{aligned}
\end{equation}
\begin{equation}
    \begin{aligned}
        \hat{\mathbf{v}}, \hat{\mathbf{v}}' &= \mathrm{Ground}(\mathbf{x},\theta_\psi), \mathrm{Ground}(\mathbf{x}',\theta_\psi), 
    \end{aligned}
\end{equation}
\begin{equation}
    \label{eqn:zero-loss}
    \begin{aligned}
    \mathcal{L}_\mathrm{zero}
    = & \mathrm{Div}_\mathrm{JS}\!\Big(\hat{\mathbf{v}} \odot \mathbb{I}\big(\bm{t}^{\psi, \underline{\mathtt{C}}_i} = 0\big) \,\Big\|\, \hat{\mathbf{v}}' \odot \mathbb{I}\big(\bm{t}^{\psi, \underline{\mathtt{C}}_i}= 0\big)\Big),
    \end{aligned}
\end{equation}
\begin{equation}
    \label{eqn:non-zero-loss}
    \begin{aligned}
    \mathcal{L}_\mathrm{one}
    =  &
    \Big(\mathrm{BCE}\big([\hat{{v}}_p, \hat{{v}}'_p], [\tfrac{1 - \delta^{\psi}_{i}}{2}, \tfrac{1 + \delta^{\psi}_{i}}{2}]\Big) * \mathrm{abs}(\delta^{\psi}_{i}),
    \end{aligned}
\end{equation}
where $\hat{\mathbf{v}}, \hat{\mathbf{v}}'\in[0,1]^P$ are the predicted ground predicates by applying $\theta_\psi$ on all possible object sets from $\mathcal{O}$.
$\mathrm{Div}_\mathrm{JS}(\cdot | \cdot)$ denotes the Jensen–Shannon divergence~\cite{lin1991divergence} and Eq.\eqref{eqn:zero-loss} tries to minimize the distance between $\hat{\mathbf{v}}$ and $\hat{\mathbf{v}}'$ if the indices with zero values in $\bm{t}^{\psi, \underline{\mathtt{C}}}$.
$\hat{{v}}_p,\hat{{v}}_p'$ denotes $p$-th ground predicate, where $\mathcal{O}_{\underline{\psi}_p} \subseteq \mathcal{O}_{\underline{\mathtt{C}}_i}$. $\mathrm{BCE}(\cdot,\cdot)$ represents the Binary Cross-Entropy, which tries to directly supervise $\hat{\mathbf{v}}$ and $\hat{\mathbf{v}}'$ if $\delta^{\psi}_{i}\neq 0$.
Intuitively, $\mathcal{L}_\mathrm{zero}$ supervises the ground predicates whose values should stay unchanged in a transition (but we don't know their values).
$\mathcal{L}_\mathrm{one}$, on the other hand, supervises the ground predicates whose values can be derived based on the effect vectors.
Since we have $\Delta^\psi$ for all lifted actions, the pipeline can be applied to all ground transition pairs in $\mathcal{D}$, resulting in the global loss,
\begin{equation}
\label{eqn:global_loss}
    \mathcal{L}^{\mathcal{D}}(\theta_\psi)
    = \sum_{\mathtt{C}\in\mathcal{C}} \mathbb{E}_{(\mathbf{x},\underline{\mathtt{C}},\mathbf{x'})\sim\mathcal{D}_\mathtt{C}}\mathcal{L}(\mathbf{x}, \mathbf{x}', \theta_\psi),
\end{equation}
where $\mathcal{D}_\mathtt{C}$ denotes the distribution of the grounded transition for action $\mathtt{C}$ in the dataset $\mathcal{D}$ (See \fref{fig:method_1} for the examples of two transitions from different actions).
Since $\mathcal{L}$ is fully differentiable with respect to $\theta$, given a effector $\Delta^{\psi}$ and the demonstration dataset, we could leverage the general and standard deep learning frameworks~\cite{kingma2014adam,rumelhart1986sgd} to train a neural classifier $\theta$ that minimizes the loss $\mathcal{L}$ for any state representation.

\subsection{Neural Loss as Guidance for Symbolic Learning}

To obtain all the classifiers $\theta_{\Psi^{\mathtt{Var}}}$ for all the typed predicates $\psi\in\Psi^{\mathtt{Var}}$, the problem now becomes finding all the lifted effect vectors $\Delta^\psi\in\Delta^{\Psi^{\mathtt{Var}}}$.
As defined in \defref{def:effect_vec}, the lifted effect vectors live in the discrete world with a finite shape, which motivates us to establish some discrete optimization strategies for it.
One insight here is that, despite the large space, only a few effect vectors provide reasonable supervision, with unreasonable ones resulting in high classification error on the validation set after the training converges.

A motivating example is depicted in \fref{fig:method_1}, where the proposed effect vector assumes the predicate to be the add effects for both $\mathtt{Gaze}$ and $\mathtt{Grasp}$.
In the demonstration, $\mathtt{Grasp}$ closely follows $\mathtt{Gaze}$, making the intermediate state shared in the two transition pairs but with one being the next state and the other being the current state.
Then, the intermediate state will be labeled as both $\mathtt{True}$ and $\mathtt{False}$, which is unreasonable and results in high classification error.
Thus, our symbolic learning aims to efficiently find all ``reasonable" effect vectors,
\begin{definition}[Symbolic Learning Objective]
    Let the demonstrations be split into non-overlapping training and validation sets $\mathcal{D} = \mathcal{D}^\text{train} \cup \mathcal{D}^\text{val}$, the objective of our symbolic learning is to find a subset of effect vectors $\Delta^{*,\Psi^{\mathtt{Var}}} \subset \Delta^{\Psi^{\mathtt{Var}}}$, 
    \begin{equation}\label{eqn:search_obj}
        \Delta^{*,\Psi^{\mathtt{Var}}} = \left\{ \Delta^\psi \in \Delta^{\Psi^{\mathtt{Var}}} \mid \mathcal{L}^{\mathcal{D}^\text{val}}(\theta_\psi) \leq \tau \right\},
    \end{equation}
    where $\theta^\psi$ is learned from $\mathcal{D}^\text{train}$ with supervision derived from $\Delta^\psi$ using Eq.~\eqref{eqn:global_loss}, and $\mathcal{L}^{\mathcal{D}^\text{val}}$ denotes the validation loss of classifier $\theta^\psi$ calculated from Eq.~\eqref{eqn:global_loss}, and $\tau$ is a given threshold.
\end{definition}

\begin{figure}[!t]
	\centering
	\includegraphics[width=0.9\columnwidth]{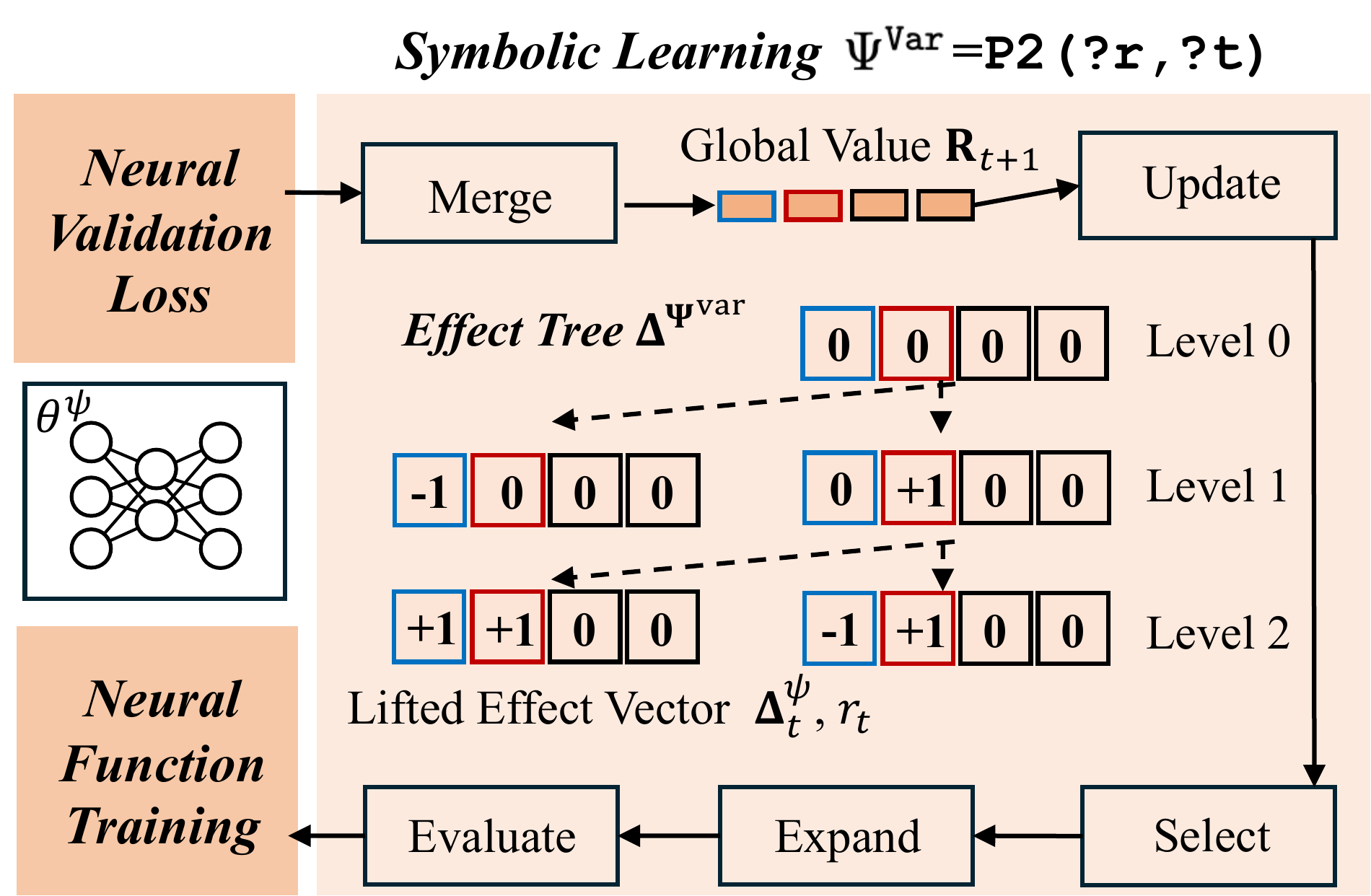}
	\caption{Detailed symbolic learning process for predicate group $\texttt{P2(?r,?t)}$. With the neural validation loss from the previous iteration, symbolic learning starts by merging the loss into the global value vector (See Eq. (7)). After the node values in the effect tree are updated, we conduct parent node selection and expansion.
    Among the children nodes, we evaluate the child with the current highest value, via the neural classifer training process described in \fref{fig:method_1}.
    }
	\label{fig:method_2}
\end{figure}

Inspired by the fact that a predicate's effects are usually sparse among actions, we propose a tree expansion algorithm for efficiently learning $\Delta^{*,\Psi^{\mathtt{Var}}}$.
Specifically, as shown in \fref{fig:method_2} (with $M=4$ actions), the complete effect vector set, $\Delta^{\Psi^{\mathtt{Var}}}$, is formulated into a tree-like structure, with each node $\Delta^{\psi_n}\in\Delta^{\Psi^{\mathtt{Var}}}, n>0$ representing an effect vector.
The root $\Delta^{0}$ of the tree is an ``all-zero" effect vector, which is not associated with any potential \textit{dynamic} predicate.
The nodes in the $l-$th level represent a vector with $l$ non-zero entries.
For each non-leaf node in the $l-$th level, its ``children" in the $l+1$-th level have the same non-zero entries with one more non-zero entry.
A naive exploration in this tree is to enumerate its nodes and train neural classifiers with supervisions from each of them, which is extremely time-consuming due to the large space.
To explore the effect tree more efficiently, \model{} tries to balance the trade-off between \textit{exploration} and \textit{exploitation}~\cite{coulom2006mcts,silver2017alphago}.

Specifically, each node $\Delta^{\psi_n}$ in the tree is additionally associated with a scalar $r^n_t$, indicating its value for finding $\Delta^\psi \in\Delta^{*,\Psi^{\mathtt{Var}}}$ if we expand it at the current $t$-th iteration.
In the $t$-th iteration, we start by selecting a parent node $\mathrm{Par}(\Delta^{\psi}_t)$ with the highest current Upper Confidence bounds applied to Trees (UCT) score~\cite{silver2017alphago}.
Its child, $\Delta^{\psi}_t$ that has the current highest value $r_t$ among all the children is proposed for evaluation (index is neglected for simplicity).
The evaluation process is defined as the supervised neural learning process in \sref{sec:neural_learning}.
After the classifier $\theta^{\psi}$ converges, we collect its \textit{action-wise} validation loss $\mathbf{L}_t\in\mathbb{R}^M$ by decomposing Eq.~\eqref{eqn:global_loss} for each action $\mathtt{C}\in\mathcal{C}, |\mathcal{C}|= M$.
The loss information is then used to update the values of all the nodes in the tree, which helps us select and expand the parent node in the $t+1$-th iteration.
The tree expansion terminates if all of the existing nodes are fully expanded, or, if the max iteration has been reached.

\begin{figure*}[!t]
	\centering
	\includegraphics[width=1.8\columnwidth]{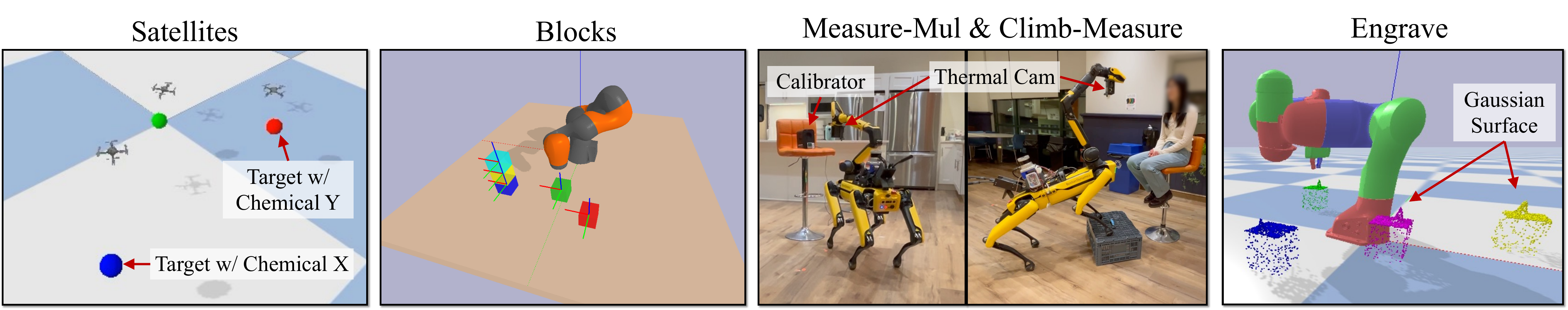}
	\caption{Visualization of the five domains (excluding Climb-Transport) we have studied in this work. These domains feature various state representations (including the high-dimensional point clouds in the Engrave domain) where our \model{} can be generally applied.}
	\label{fig:domains}
\end{figure*}

\begin{table*}[!t]
    \centering
    \setlength{\tabcolsep}{0.9mm}
    \fontsize{8}{10}\selectfont
    \begin{tabular}{cccc|ccc|ccc|ccc|ccc|ccc}
    \toprule[1.5pt]
    Domain & \multicolumn{3}{c}{Satellites~\cite{kumar2023predict}} & \multicolumn{3}{c}{Blocks~\cite{chitnis2021nsrt}} & \multicolumn{3}{c}{Measure-Mul} & \multicolumn{3}{c}{Climb-Measure} & \multicolumn{3}{c}{Climb-Transport} & \multicolumn{3}{c}{Engrave} \\
    State Space & \multicolumn{3}{c}{SE2 ($\mathbb{R}^{3\times5}$)} & \multicolumn{3}{c}{Vec3 ($\mathbb{R}^{3\times8}$)} & \multicolumn{3}{c}{SE3 ($\mathbb{R}^{7\times5}$)} & \multicolumn{3}{c}{SE3 ($\mathbb{R}^{7\times5}$)} & \multicolumn{3}{c}{SE3 ($\mathbb{R}^{7\times5}$)} & \multicolumn{3}{c}{PCD ($\mathbb{R}^{1024\times3\times6}$)} \\
    \midrule
    Test Dist & $\mathcal{T}^\mathrm{train}$    & $\mathcal{T}^\mathrm{test}$   &  $\downarrow$ & $\mathcal{T}^\mathrm{train}$    & $\mathcal{T}^\mathrm{test}$   &  $\downarrow$ & $\mathcal{T}^\mathrm{train}$    & $\mathcal{T}^\mathrm{test}$   &  $\downarrow$ & $\mathcal{T}^\mathrm{train}$    & $\mathcal{T}^\mathrm{test}$   &  $\downarrow$ & $\mathcal{T}^\mathrm{train}$    & $\mathcal{T}^\mathrm{test}$   &  $\downarrow$ & $\mathcal{T}^\mathrm{train}$    & $\mathcal{T}^\mathrm{test}$   &  $\downarrow$ \\
    Oracle & 100.0 & 100.0 & 0.00  & 100.0 & 100.0 & 0.00  & 100.0 & 100.0 & 0.00  & 90.0  & 81.6  & 0.09  & 91.2  & 82.0  & 0.10  & 100.0 & 100.0 & 0.00 \\
    \textbf{IVNTR (Ours)} & \underline{\textbf{99.2}}  & \underline{\textbf{93.2}}  & \underline{\textbf{0.06}}  & \underline{\textbf{100.0}} & \underline{\textbf{82.0}}  & \underline{\textbf{0.18}}  & \underline{\textbf{90.0}}  & \underline{\textbf{88.4}}  & \underline{\textbf{0.02}}  & \underline{\textbf{91.6}}  & \underline{\textbf{65.6}}  & \underline{\textbf{0.28}}  & \underline{\textbf{79.2}}  & \underline{\textbf{53.2}}  & \underline{\textbf{0.33}}  & \underline{\textbf{98.4}}  & \underline{\textbf{79.2}}  & \underline{\textbf{0.20}} \\
    GNN~\cite{battaglia2018gnn}   & 74.0  & 6.0   & 0.92  & 82.4  & 24.0  & 0.71  & 2.4   & 1.2   & 0.50  & 20.8  & 0.7   & 0.97  & 48.0  & 2.0   & 0.96  & 0.0   & 0.0   & 1.00 \\
    Transformer~\cite{vaswani2017tf} & 46.8  & 1.2   & 0.97  & 24.4  & 7.6   & 0.69  & 10.0  & 2.4   & 0.76  & 29.2  & 0.0   & 1.00  & 10.4  & 0.8   & 0.92  & 0.0   & 0.0   & 1.00 \\
    FOSAE~\cite{asai2019latplan_fol} & 100.0 & 34.8  & 0.65  & 2.4   & 0.4   & 0.83  & 3.6   & 1.2   & 0.67  & 21.2  & 0.0   & 1.00  & 45.6  & 0.8   & 0.98  & 0.0   & 0.0   & 1.00 \\
    Grammar~\cite{silver2023predicateinvent} & 0.0   & 0.0   & 1.00  & 0.0   & 0.0   & 1.00  & 0.0   & 0.0   & 1.00  & 0.0   & 0.0   & 1.00  & 0.0   & 0.0   & 1.00  & N/A   & N/A   & N/A \\
    Random & 0.0   & 0.0   & 1.00  & 10.6  & 1.2   & 0.89  & 0.0   & 0.0   & 1.00  & 0.0   & 0.0   & 1.00  & 0.0   & 0.0   & 1.00  & 0.0   & 0.0   & 1.00 \\
    \bottomrule[1.5pt]
    \end{tabular}%
  \caption{Empirical success rate comparison on six simulated robot planning domains.
  Among all the methods, \model{} suffers the least from the generalization challenge due to its relational structure.
  The scores are obtained from the averaged results over five random seeds.
  }
  \label{tab:sim_emp}%
\end{table*}%

Clearly, the key to more efficient learning lies in the definition and updating strategy of the node values $r^n_t$.
As the sparsity of predicate effects among actions is indicated by the sparsity of non-zero entries in the effect vector, we try to efficiently explore the entries in the effect vectors that \textbf{should not} be zero to optimize Eq.~\eqref{eqn:search_obj}.
To achieve this, we try to leverage the guidance from the ``zero-parts" (Eq.~\eqref{eqn:zero-loss}) of $\mathbf{L}_t$.
Specifically, after the evaluation of the node $\Delta^{\psi}_t$ in the $t-$th iteration with $\mathbf{L}_t$, we use the following equations to update and compute $r^n_{t+1}$ for all the nodes in the tree,
\begin{equation}
\label{eqn:update}
    \begin{aligned}
    \mathbf{R}_{t+1} &= 
    \mathbf{R}_{t} \odot \mathbb{I}(\Delta^{\psi}_t \neq 0) 
    + \frac{(\mathbf{R}_{t} + \mathbf{L}_t)}{2} \odot \mathbb{I}(\Delta^{\psi}_t = 0). \\
        r^n_{t+1} &= \mathrm{Mean}\big(\mathbf{R}_{t+1} \odot \mathbb{I}(\Delta^{\psi_n} = 0)\big),
    \end{aligned}
\end{equation}
where $\mathbf{R}_{t}\in\mathbb{R}^M, \mathbf{R}_0=\mathbf{0}^M$ is a global value vector that stores the information from historical evaluations.
$\mathbb{I}(\Delta^{\psi}_t = 0)$ is a binary vector indicating if an entry in the evaluated node $\Delta^{\psi}_t$ equals to zero.
Here, we only update $\mathbf{R}_{t}$ with the ``zero-parts" of the loss information $\mathbf{L}_t$, which then helps update the node values.
The node values intuitively indicate how likely the loss will decrease in its children where
there are fewer zeros in the effect vectors.
From Eq.~\eqref{eqn:update}, we see that the higher loss in the zero entry indexes of $\Delta^{\psi_n}$ contributes to its higher value, encouraging the evaluation to prioritize its children, which are likely to decrease the loss.
In \appref{app:pruning}, we additionally introduce some pruning strategy for more efficient expansion.

Finally, we collect all the effect vectors $\Delta^{\psi_n}\in\Delta^{*,\Psi^{\mathtt{Var}}}$ with the associated neural classifier $\theta_{\psi_n}$ as the outcomes from the bilevel learning of typed predicates $\Psi^{\mathtt{Var}}$.
As shown in \fref{fig:overview} (a), there could exist multiple different vector-classifier pairs for the same typed predicate $\Psi^{\mathtt{Var}}$, which are treated as different predicates in the following predicate selection stage.\footnote{Following previous work~\cite{silver2023predicateinvent}, we can also add quantifiers and negations as prefix for each of the invented neural predicates.}

\subsection{Predicate Selection}
With all possible typed predicates, \model{} is able to construct the predicate pool $\hat{\Psi}_{\mathrm{dyn}}$ without any classifier pre-definition.
This strength has made bilevel planning applicable to complicated and high-dimensional state spaces.
Meanwhile, as the predicate classifiers are \textit{relational} and can be seamlessly integrated into an AI planner~\cite{helmert2006fast}, our approach can naturally achieve compositional generalization.

Yet, not all of the predicates in $\hat{\Psi}_{\mathrm{dyn}}$ are favorable for planning.
Therefore, we next try to select a subset $\tilde{\Psi}_{\mathrm{dyn}}\subset\hat{\Psi}_{\mathrm{dyn}}$ that minimizes the score function $J(\tilde{\Psi}_{\mathrm{dyn}},
\mathcal{D})$~\cite{silver2023predicateinvent}.
Specifically, with a set of candidate predicates $\tilde{\Psi} = \{\Psi_\mathrm{G},\Psi_\mathrm{sta}, \tilde{\Psi}_{\mathrm{dyn}}\}$, we start by grounding all the states $\mathbf{x}$ in the demonstration $\mathcal{D}$, which forms a ground atom dataset.
Since we already have the effect vector for each of the predicates in $\tilde{\Psi}_{\mathrm{dyn}}$, the effect sets ($\tilde{\mathtt{Eff}}^+, \tilde{\mathtt{Eff}}^-\subset \tilde{\Psi}_{\mathrm{dyn}}$) for each operator $\tilde{\mathtt{Op}}^\mathtt{C}$ can be easily obtained.
Then, the precondition set for each operator can be calculated using an intersection strategy~\cite{chitnis2021nsrt}.
The learned operator set $\tilde{\mathtt{Op}}$ is then applied to the ground atom dataset to generate plan \textit{skeletons} $\tilde{\pi}$ for each task, which are compared with the demonstration plan \textit{skeletons} $\bar{\pi}$ for objective calculation~\cite{silver2023predicateinvent}.
The objective is finally minimized by running a hill-climbing search over $\hat{\Psi}_{\mathrm{dyn}}$, resulting in the desired predicate set ${\Psi}_{\mathrm{dyn}}$.
With the complete set, we are now able to learn the planning \textit{abstractions} (operators and samplers) using standard pipelines~\cite{chitnis2021nsrt,kumar2023predict,liang2024visualpredicator} as shown in \fref{fig:overview} (a).
For a more detailed explanation to operator and sampler learning, please see Appendix \appref{app:op_sam_learning}.

Note that before the predicate selection stage, all of the predicates' neural classifiers have been pre-trained using our \model{} framework, which has avoided the challenge of using the planning objective for learning but still achieved a powerful and adaptive model optimized for planning.

\section{Experiments}

\begin{table*}[!t]
    \centering
    \setlength{\tabcolsep}{1.5mm}
    \fontsize{8}{10}\selectfont
    \begin{tabular}{cccccccccc|cccccccc}
    \toprule[1.5pt]
          &       & \multicolumn{8}{c}{Climb-Measure}                             & \multicolumn{7}{c}{Climb-Transport}                       &  \\
    Planner & Seed/Task & T0    & T1    & T2    & T3    & T4    & T5    & Mean  & Avg.  & T0    & T1    & T2    & T3    & T4    & T5    & Mean  & Avg. \\
    \midrule
    \midrule
    \multirow{3}[2]{*}{Oracle (Human)} & S0    & 0.0     & 1.0     & 1.0     & 1.0     & 0.5   & 1.0     & 0.750 & \multirow{3}[2]{*}{0.833} & 0.5   & 0.0     & 1.0     & 1.0     & 1.0     & 1.0     & 0.750 & \multirow{3}[2]{*}{0.592} \\
          & S1    & 1.0     & 1.0     & 1.0     & 0.5   & 1.0     & 0.5   & 0.833 &       & 1.0    & 1.0    & 0.33  & 0.5   & 0.5   & 0.5   & 0.638 &  \\
          & S2    & 1.0    & 1.0    & 1.0    & 1.0    & 1.0    & 0.5   & 0.917 &       & 0.5   & 0.33  & 0.5   & 0     & 0.5   & 0.5   & 0.388 &  \\
    \midrule
    \multirow{3}[2]{*}{\textbf{IVNTR (Learned)}} & S0    & 1.0   & 1.0   & 0.5   & 0.0   & 1.0   & 1.0   & 0.750 & \multirow{3}[2]{*}{0.778} & 0.5   & 1.0    & 0     & 0.5   & 1.0    & 0.33  & 0.555 & \multirow{3}[2]{*}{0.546} \\
          & S1    & 1.0   & 1.0   & 1.0   & 1.0   & 1.0   & 0.0   & 0.833 &       & 0.33  & 0.5   & 1.0    & 0.33  & 0     & 0.5   & 0.443 &  \\
          & S2    & 0.5   & 1.0   & 1.0   & 1.0   & 0.0   & 1.0   & 0.750 &       & 0.5   & 0.5   & 1.0    & 1.0    & 0.5   & 0.33  & 0.638 &  \\
    \bottomrule[1.5pt]
    \end{tabular}%
  \caption{Success rate comparison of the two planners on the real robot planning tasks sampled from $\mathcal{T}^\mathrm{test}$. 
  ``Oracle" denotes bilevel planners exhaustively engineered by a human expert. 
  \model{} is learned from the demonstrations collected in the simulated environment.
  For each domain, we have tested six tasks (T0$\sim$T5) sampled from three random seeds (S0$\sim$S2).
  For each planner in each task, we run it at most three times and record the averaged task success rate.
  Our framework has achieved comparable real robot performance with the Oracle. 
  }
  \label{tab:real_emp}%
\end{table*}%

\begin{table}[!t]
\centering
\setlength{\tabcolsep}{1.0mm}
    \fontsize{8}{10}\selectfont
    \begin{tabular}{ccccccc}
        \toprule[1.5pt]
        Domains & \multicolumn{3}{c}{Blocks} & \multicolumn{3}{c}{Climb-Measure} \\
        Metric & $\mathcal{T}^\mathrm{train}$    & $\mathcal{T}^\mathrm{test}$   &  $J(\cdot)$ ($\times10^5$) & $\mathcal{T}^\mathrm{train}$    & $\mathcal{T}^\mathrm{test}$   & $J(\cdot)$ ($\times10^5$) \\
        \midrule
        GT-Vectors & 80.0  & 62.8  & 121.59 & 90.4  & 61.2  & 2.718 \\
        \textbf{IVNTR} & \textbf{100.0} & \textbf{82.0}  & \textbf{56.77} & \textbf{91.6}  & \textbf{65.6}  & \textbf{2.481} \\
        \bottomrule[1.5pt]
    \end{tabular}%
    \caption{Comparison between the predicates learned with ground-truth (GT)-vectors and using our \model{} framework. We report the task success rate and the final planning objective.}
  \label{tab:abla_gt_vec}%
\end{table}%

\subsection{Implementation Details}
\myparagraph{System and Hardware:} All methods are evaluated on a single NVIDIA A100 GPU and an AMD EPYC 7543 32-Core CPU to ensure fairness. 
Training is conducted on the same hardware as evaluation, with domain-specific details provided in \appref{app:domain_details}. 
Real-robot experiments are performed using the Boston Dynamics Spot robot equipped with an arm.

\myparagraph{Baselines:} Since we do not assume access to the complete predicate set, most existing bilevel planning approaches~\cite{chitnis2021nsrt,kumar2023predict,kumar2024practice} are inapplicable. 
We attempted the grammar-based approach~\cite{silver2023predicateinvent}, but it failed to optimize the planning objective in most domains (see \appref{app:objective}). 
Thus, \model{} is primarily compared to relational neural policy learning methods~\cite{battaglia2018gnn,vaswani2017tf,asai2019latplan_fol}. 
Following prior works~\cite{kumar2023predict,chitnis2021nsrt,silver2023predicateinvent,silver2022skills}, baselines are trained using standard behavior cloning pipelines and evaluated with a shooting strategy; see \appref{app:baseline_details} for details.

\myparagraph{Domains:}
We evaluate the methods across six diverse robot planning domains with varying state representations, as visualized in \fref{fig:domains} and summarized in \tref{tab:sim_emp}. 
Below, we provide a high-level overview of these domains.
For more implementation details, please refer to \appref{app:domain_details}:
\begin{itemize}
    \item \textit{Satellites} comes from prior work~\cite{kumar2023predict}, which involves a group of satellites collaborating to capture sensor readings from targets. 
    States comprise SE2 poses and object attributes. 
    Training scenarios ($\mathcal{T}^\mathrm{train}$) include 2 satellites and 2 targets, while test scenarios ($\mathcal{T}^\mathrm{test}$) have 3 of each.
    \item \textit{Blocks:} Inspired by~\cite{chitnis2021nsrt}, this domain tasks a robot with manipulating 3D blocks to form goal towers. 
    Unlike vanilla Blocks World, the goals here involve ``packing" pairs of blocks into two-level towers.
    $\mathcal{T}^\mathrm{train}$ includes 4–5 blocks, while $\mathcal{T}^\mathrm{test}$ features 6–7 blocks.
    \item \textit{Measure-Mul:} In this new domain inspired by Satellites, a Spot robot calibrates a thermal camera by aligning it with a calibrator before measuring body temperatures of multiple human targets. 
    States include 6D poses of the robot and the targets. 
    Training distributions ($\mathcal{T}^\mathrm{train}$) have 2–3 humans, while test distributions ($\mathcal{T}^\mathrm{test}$) include 4.
    \item \textit{Climb-Measure} is similar to Measure-Mul but with added complexity: calibrators and human targets may be placed at high, initially unreachable locations. 
    The Spot robot must arrange platforms and climb onto them to reach targets. 
    Training ($\mathcal{T}^\mathrm{train}$) includes 0–1 platforms, while testing ($\mathcal{T}^\mathrm{test}$) requires planning with 2 platforms.
    \item \textit{Climb-Transport:} Introduced in \fref{fig:running_example}, this domain requires the Spot robot to arrange platforms to grasp a high-placed target, then transport it into a container. 
    Training setups ($\mathcal{T}^\mathrm{train}$) feature 0–1 platforms, while testing ($\mathcal{T}^\mathrm{test}$) involves 2 platforms.
    \item \textit{Engrave} features high-dimensional state spaces represented as object-centric point clouds. 
    Similar to Blocks, the goal is to ``pack" blocks. 
    However, blocks start with one irregular Gaussian surface that must be ``engraved" and ``rotated" to create a matching fit. 
    Training distributions ($\mathcal{T}^\mathrm{train}$) include 3–4 blocks, while testing ($\mathcal{T}^\mathrm{test}$) has 5–6.
\end{itemize}

\myparagraph{Experiment Setup:}
For each domain, we manually designed an oracle bilevel planner (Oracle) to collect training demonstrations. 
We report averaged results over five random seeds for all six domains.
For each seed in Satellites, Blocks, Measure-Mul, and Engrave, we have collected $500$ demonstrations.
For Climb-Measure and Climb-Transport, $2000$ demonstrations were collected per seed.
During test, each seed in each domain includes $50$ in-domain tasks ($T\sim\mathcal{T}^\mathrm{train}$) and $50$ generalization tasks ($T\sim\mathcal{T}^\mathrm{test}$). 
We report the success rate within the same maximum planning time for all the methods.
For real-world Climb-Measure and Climb-Transport, a shared map was recorded using Spot’s default graph\_nav service for simulation and localization. 
Each domain was tested on $3$ random seeds, each with $6$ generalized tasks. 
For manipulation-based actions, we have utilized an off-the-shelf segment anything model (SAM)~\cite{lang_sam,ravi2024sam} for computing the grasping pixel.

\begin{table}[!t]
\centering
\setlength{\tabcolsep}{2mm}
    \fontsize{8}{10}\selectfont
    \begin{tabular}{ccccccc}
        \toprule[1.5pt]
        Domains & \multicolumn{3}{c}{Satellites} & \multicolumn{3}{c}{Measure-Mul} \\
        Sampling with $\theta^\psi$ & \checkmark    & \xmark   &  $\downarrow$ & \checkmark    & \xmark   & $\downarrow$ \\
        \midrule
        $\mathcal{T}^\mathrm{train}$ & \textbf{99.2}  & 74.0  & 0.254 & \textbf{90.0}  & 2.8   & 0.969 \\
        $\mathcal{T}^\mathrm{test}$ & \textbf{93.2}  & 16.8  & 0.820 & \textbf{88.4}  & 1.4   & 0.984 \\
        \bottomrule[1.5pt]
    \end{tabular}%
    \caption{Comparison between sampling with and without the invented predicate classifiers. Without using the predicates as step-wise success indicator, the performance drops significantly.}
  \label{tab:abla_highlevel}%
\end{table}%

\subsection{Empirical Results}
\myparagraph{Simulated Planning Domains:}
The empirical comparison across the six simulated domains is presented in \tref{tab:sim_emp}. 
Alongside the averaged success rates, we report the performance drop percentage during generalization.
\model{} consistently outperforms all baselines in both $\mathcal{T}^\mathrm{train}$ and $\mathcal{T}^\mathrm{test}$ across all domains. 
For complex state representations such as SE3 and PointClouds (PCD), none of the baselines achieve a success rate above $5\%$ on generalized tasks, while \model{} stably solves over $50\%$ by virtue of its relational structure.
In \appref{app:blocks_img}, we further demonstrate the potential of \model{} to abstract high-dimensional RGB image states into symbolic predicates.

\myparagraph{Real Robot Planning Tasks:}
All real robot tasks are sampled from $\mathcal{T}^\mathrm{test}$, making \model{} the only applicable approach. 
To benchmark performance, a human expert has exhaustively engineered oracle planners for the real robot in the two domains. 
Each approach attempts each task up to three times, with the average success rate reported in \tref{tab:real_emp}.
Despite the simulation-to-reality gap, \model{} successfully generalizes to held-out tasks, achieving results comparable to the oracle planner. 
Most real-world deployment failures stem from perception and localization errors, with examples shown in \appref{app:failures}.

\subsection{Ablation Studies}
\myparagraph{Comparison to Ground-Truth Effect Vectors:}
A notable strength of \model{} is its ability to discover non-ground-truth (GT) predicates. 
In \tref{tab:abla_gt_vec}, we replaced our tree expansion with oracle-derived GT effect vectors, where the performance on the Blocks and Climb-Measure domains are reported. 
Interestingly, \model{} minimizes the planning objective more effectively than GT vectors, resulting in its higher accuracy. 
This outcome highlights the advantage of exploring better high-level abstractions beyond human-engineered ones.

\begin{figure}[!t]
	\centering
	\includegraphics[width=1.01\columnwidth]{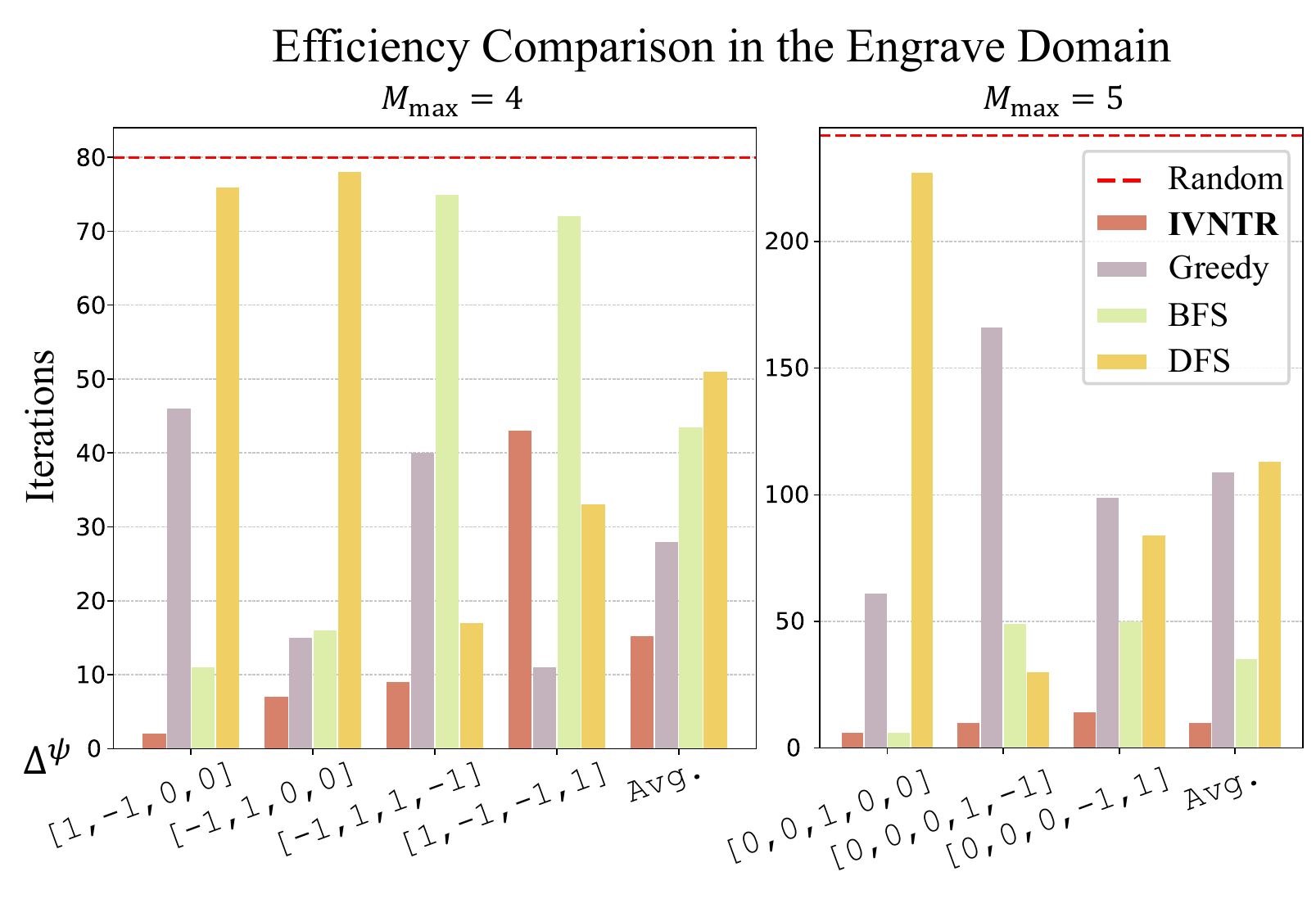}
	\caption{Comparison between \model{} with other search strategies in the Engrave domain. We report the number of iterations for each of the algorithm to find the reasonable effect vectors (different predicate could have different maximum search space $M_{\mathrm{max}}$). \model{} has demonstrated the highest efficiency in finding the desired vectors.}
	\label{fig:abla_search}
\end{figure}

\myparagraph{Comparison to Other Search Algorithms:}
To evaluate the efficiency of our neural-informed tree expansion algorithm, we compared it with alternative search strategies: a greedy approach (Greedy) that flips the zero entry with the highest current loss, breadth-first (BFS) and depth-first (DFS) searches.
For the Engrave domain, \fref{fig:abla_search} shows the number of iterations required to find each reasonable effect vector using these methods. 
Compared to uninformed methods, our \model{} framework is at least $2\times$ more efficient.
The greedy approach exhibits high variance and is generally less reliable. 
We present comparisons in more domains in \appref{app:more_search}.

\myparagraph{Comparison to Pure High-level Planning:}\label{sec:abla_pure_high_level}
As discussed in \sref{sec:bilevel_planning}, predicates not only enable compositional generalization through planning operators but also serve as indicators of sampler failures in low-level states. 
To assess the importance of our invented predicates for reliable low-level sampling, we disabled bilevel planning and followed the high-level plan greedily, ignoring predicate-based checks for sampler success.
As shown in \tref{tab:abla_highlevel}, this approach results in performance drops of up to $98.4\%$, underscoring the critical role of predicates as indicators of low-level state validity.

\subsection{Interpreting Invented Predicates}

Predicates play a key role in defining the preconditions and effects of operators, specifying the order of ground actions to complete a task. 
In \tref{tab:interprete}, we display part of the precondition and effect sets for high-level actions in the Climb-Transport domain, where invented predicates capture logical relationships among actions.
For example, the $\mathtt{Drop}$ action requires $\mathtt{P2_1}$ and $\mathtt{P2_2}$, which are the add effects of $\mathtt{Gaze}$, $\mathtt{MTGaze}$, and $\mathtt{WalkOn}$. 
Similarly, the $\mathtt{Pick}$ action depends on $\mathtt{P1_1}$, the delete effect of $\mathtt{Grasp}$, enforcing all $\mathtt{Pick}$ actions to precede $\mathtt{Grasp}$. 
These relational constraints over objects enable the generation of long-horizon plans that generalize to unseen object compositions. 
The complete operators are detailed in \appref{app:complete_op}.
We also provide visualizations of how predicates act as success indicators to filter out sampler failures in \appref{app:sampler_vis}.

\section{Related Works}

\subsection{Learning Abstractions for Planning}
Learning abstractions is essential for reducing the complexity of long-horizon planning in high-dimensional domains.
Traditional approaches, such as hierarchical task networks (HTN)~\cite{kaelbling2011htn}, heavily rely on hand-designed abstractions.
Recent advances have shifted towards data-driven approaches that automatically discover useful abstractions from interactions~\cite{gupta2020relay,Soroush2022maple,hansen2022bisimulation,dong2019nlm,chitnis2021glib} or demonstrations~\cite{sharmadirected,kipf2019compile,chitnis2021nsrt,mao2022pdsketch}.
However, these methods struggle to generalize beyond the training environments~\cite{liang2024visualpredicator}.
Foundation models, such as large language models (LLMs) and vision-language models (VLMs), have been explored for high-level planning with minimal or no demonstrations~\cite{liu2024BLADE,fang2024keypoint,liang2024visualpredicator,silver2024generalized,wei2022cot,han2024interpret,huang2023voxposer,hu2023look,kumar2024openworld}. 
While these models leverage commonsense knowledge for efficient plan generation, two challenges remain:
(1) High-level plans from LLMs~\cite{silver2024generalized,han2024interpret,wei2022cot,huang2023voxposer} are difficult to reliably refine in the low-level space~\cite{liang2024visualpredicator}.
(2) VLM-based methods~\cite{liang2024visualpredicator,fang2024keypoint,kumar2024openworld,yang2024guidinglonghorizontaskmotion} struggle in domains where images cannot fully capture the state space.

\subsection{Task and Motion Planning}
To address these challenges, task and motion planning (TAMP) integrates high-level symbolic planning with low-level motion generation. 
Traditional TAMP methods~\cite{garrett2021integrated,garrett2020pddlstream} rely on manually designed planners~\cite{McDermott1998PDDL,karaman2011anytime}. 
These frameworks inherently support compositional generalization due to their relational structure. 
In addition, the coupling between high-level and low-level planning can handle failures at either level effectively.
However, traditional TAMP requires substantial human effort. 
Recent advances integrate learning into TAMP~\cite{chitnis2021nsrt,bougie2020skill,kumar2023predict,silver2023predicateinvent,liang2024visualpredicator,kumar2024openworld,yang2024guidinglonghorizontaskmotion}, forming bilevel planning frameworks. 
These approaches combine the strengths of TAMP with the scalability of machine learning models. 
However, most bilevel planners rely on manually engineered state abstractions (predicates), limiting their scalability and flexibility.

\begin{table}[!t]
    \centering
    \setlength{\tabcolsep}{1.2mm}
    \fontsize{8}{10}\selectfont
    \begin{tabular}{cccccc}
    \toprule[1.5pt]
     Predicates & $\mathtt{P1_1(?r)}$ & $\mathtt{P2_1(?r,?t)}$ & $\mathtt{P2_2(?r,?t)}$ & $\mathtt{In(?t,?t)}$ \\
    \midrule
    \texttt{Grasp}      & $\mathtt{Pre} \mid \mathtt{Eff}^-$ & $\mathtt{Pre}$   & $\mathtt{Pre}$   &       \\
    \texttt{Gaze}       &       & $\mathtt{Pre}$   & $\mathtt{Eff}^+$   &       \\
    \texttt{MAOff}      & $\mathtt{Pre}$   &       &       &       \\
    \texttt{MAOn}       &       & $\mathtt{Pre}$   & $\mathtt{Pre}$   &       \\
    \texttt{MTGaze}     &       & $\mathtt{Eff}^+$   &       &       \\
    \texttt{WalkOn}     &       & $\mathtt{Eff}^+$   &       &       \\
    \texttt{MTPlace}    & $\mathtt{Pre}$   &       &       &       \\
    \texttt{MTReach}    &       &       &       &       \\
    \texttt{Pick}       & $\mathtt{Pre}$   &       &       &       \\
    \texttt{Drop}       &       & $\mathtt{Pre}$ & $\mathtt{Pre}$ & $\mathtt{Eff}^+$ \\
    \bottomrule[1.5pt]
    \end{tabular}
  \caption{The preconditions, add effects, and delete effects for each of the actions (the variables are neglected for simplicity) with (part of) the invented predicates in the Climb-Transport Domain. 
  $\mathtt{MA}$ is for $\mathtt{MoveAway}$ and $\mathtt{MT}$ for $\mathtt{MoveTo}$.
  The invented predicates have specified some logical constrains over the order of actions.
  }
  \label{tab:interprete}%
\end{table}%

\subsection{Predicate Invention for Planning}
To automate predicate generation for planning, various approaches have been proposed~\cite{liang2024visualpredicator,li2023embodied,han2024interpret,silver2023predicateinvent,asai2019latplan_fol,asai2021latplanpddl,asai2018latplan_prop,hansen2022bisimulation,shah2024reals}. 
The most direct approaches~\cite{li2023embodied,han2024interpret} rely on domain knowledge, such as human-provided labels~\cite{li2023embodied} or LLM-based oracles~\cite{han2024interpret}. 
To \textit{invent} predicates, earlier approaches derive "easy" step-wise objectives, such as reconstruction~\cite{asai2018latplan_prop,asai2019latplan_fol,asai2021latplanpddl} or bisimulation~\cite{hansen2022bisimulation}.
Among these, LatPlan (FOSAE)~\cite{asai2019latplan_fol} learns relational neural abstractions for images by reconstructing states and identifying action spaces for planning, which is the closest work to \model{}. 
However, its implicit predicates are not optimized for efficient planning, limiting its applicability to domains with only nullary actions. 
Recent approaches~\cite{silver2023predicateinvent,liang2024visualpredicator} address this by learning abstractions tailored to fast planners~\cite{helmert2006fast}. 
However, these methods struggle to \textit{learn} predicate classifiers due to non-differentiable objectives. 
Consequently, they often rely on pre-defined predicate candidates from program synthesis~\cite{silver2023predicateinvent} or foundation models~\cite{liang2024visualpredicator}, which constrains their applicability in more sophisticated and high-dimensional state spaces.
Our approach is motivated by the bilevel planning framework~\cite{silver2023predicateinvent} but eliminates the need for pre-defining the candidates. 
Instead, we can learn adaptive neural classifiers for different domains, enabling more flexible and scalable learning based planning.

\section{Limitations and Future Works}
In this work, we introduced \model{}, a bilevel learning framework that invents neural classifiers as relational planning predicates. 
These predicates enable the learning of relational bilevel planners capable of generating long-horizon plans for unseen object compositions. 
At the neural level, \model{} leverages the high-level effects of predicates across actions to provide step-wise supervisions on transition pairs. 
At the symbolic level, \model{} captures the sparsity of effects through an informed tree expansion algorithm. 
By adopting neural classifiers, \model{} adapts to diverse robot planning domains with continuous and high-dimensional state representations. 
Additionally, we deployed \model{} on a mobile manipulator, demonstrating its ability to achieve compositional generalization over objects and actions in long-horizon mobile manipulation tasks.

\model{} has several limitations that we leave for future work:
(1) \model{} can only invent \textit{dynamic} predicates. 
\textit{Static} predicates are still assumed as domain-level prior knowledge.
(2) Following previous works~\cite{kumar2024practice}, we have assumed the sparsity of effects; discussions about the general cases can be found in \appref{app:assumption}.
(3) Since \model{} trains different neural networks in each iteration, the learning time could be prolonged in domains with extreme complexity.
Currently, we parallelize the invention of different predicate groups on multiple GPUs for efficiency (see \appref{app:domain_details}).
(4) Neural predicates with quantifiers are less reliable due to the prediction errors in the classifiers.
Future work could explore probabilistic symbolic planners~\cite{younes2004ppddl1} or hybrid declarative-imperative representations~\cite{mao2024hybrid} to plan with an inaccurate model.
(5) Since the predicates are neural networks, it is hard to interpret their physical meanings. It would also be intriguing for future approaches to make them directly human readable.

\section*{Acknowledgment}
We acknowledge the support of the Air Force Research Laboratory (AFRL), DARPA, under agreement number FA8750-23-2-1015.
We also acknowledge Defence Science and Technology Agency (DSTA) under contract \#DST000EC124000205.
This work used Bridges-2 at PSC through allocation cis220039p from the Advanced Cyberinfrastructure Coordination Ecosystem: Services \& Support (ACCESS) program which is supported by NSF grants \#2138259, \#2138286, \#2138307, \#2137603, and \#213296.
We express sincere gratitude to Qinglin Feng for her valuable time in supporting our real-robot experiments and for her intelligence in motivating our Climb-Transport domain.
The authors would also like to express sincere gratitude to Jiayuan Mao (MIT), Nishanth Kumar (MIT), Prof. Katia Sycara (CMU), and Prof. Pradeep Ravikumar (CMU) for their valuable feedback, discussions, and suggestions on the early stages of this work.
Finally, the authors wish to thank our Spot robot, Spotless, for being reliable throughout our real-world experiments.


\bibliographystyle{unsrt}
\bibliography{references}

\begin{thebibliography}{10}

\bibitem{mandlekar2023mimicgen}
Ajay Mandlekar, Soroush Nasiriany, Bowen Wen, Iretiayo Akinola, Yashraj Narang, Linxi Fan, Yuke Zhu, and Dieter Fox.
\newblock Mimicgen: A data generation system for scalable robot learning using human demonstrations.
\newblock {\em arXiv preprint arXiv:2310.17596}, 2023.

\bibitem{chi2023diffusionpolicy}
Cheng Chi, Siyuan Feng, Yilun Du, Zhenjia Xu, Eric Cousineau, Benjamin Burchfiel, and Shuran Song.
\newblock Diffusion policy: Visuomotor policy learning via action diffusion.
\newblock In {\em Robotics: Science and Systems (RSS)}, 2023.

\bibitem{zhao2023learning}
Tony~Z Zhao, Vikash Kumar, Sergey Levine, and Chelsea Finn.
\newblock Learning fine-grained bimanual manipulation with low-cost hardware.
\newblock In {\em Robotics: Science and Systems (RSS)}, 2023.

\bibitem{wang2024equivariant}
Dian Wang, Stephen Hart, David Surovik, Tarik Kelestemur, Haojie Huang, Haibo Zhao, Mark Yeatman, Jiuguang Wang, Robin Walters, and Robert Platt.
\newblock Equivariant diffusion policy.
\newblock {\em arXiv preprint arXiv:2407.01812}, 2024.

\bibitem{yang2024equibot}
Jingyun Yang, Zi-ang Cao, Congyue Deng, Rika Antonova, Shuran Song, and Jeannette Bohg.
\newblock Equibot: Sim (3)-equivariant diffusion policy for generalizable and data efficient learning.
\newblock {\em arXiv preprint arXiv:2407.01479}, 2024.

\bibitem{mao2024planning}
Jiayuan Mao, Tom{\'a}s Lozano-P{\'e}rez, Josh Tenenbaum, and Leslie Kaelbling.
\newblock {What Planning Problems Can A Relational Neural Network Solve?}
\newblock In {\em Proceedings of the Advances in Neural Information Processing Systems (NeurIPS)}, volume~36, 2024.

\bibitem{li2024logicity}
Bowen Li, Zhaoyu Li, Qiwei Du, Jinqi Luo, Wenshan Wang, Yaqi Xie, Simon Stepputtis, Chen Wang, Katia~P Sycara, Pradeep~Kumar Ravikumar, et~al.
\newblock {LogiCity: Advancing Neuro-Symbolic AI with Abstract Urban Simulation}.
\newblock In {\em Proceedings of the Advances in Neural Information Processing Systems (NeurIPS)}, pages 69840--69864, 2024.

\bibitem{li2006towards}
Lihong Li, Thomas~J Walsh, and Michael~L Littman.
\newblock Towards a unified theory of state abstraction for mdps.
\newblock In {\em AI\&M}, 2006.

\bibitem{abel2018state}
David Abel, Dilip Arumugam, Lucas Lehnert, and Michael Littman.
\newblock State abstractions for lifelong reinforcement learning.
\newblock In {\em International Conference on Machine Learning}. PMLR, 2018.

\bibitem{konidaris2018symbols}
George Konidaris, Leslie Pack~Kaelbling, and Tomás Lozano-Pérez.
\newblock From skills to symbols: Learning symbolic representations for abstract high-level planning.
\newblock {\em Journal of Artificial Intelligence Research}, 16:215--289, 2018.

\bibitem{wonglearning}
Lionel Wong, Jiayuan Mao, Pratyusha Sharma, Zachary~S Siegel, Jiahai Feng, Noa Korneev, Joshua~B Tenenbaum, and Jacob Andreas.
\newblock {Learning Grounded Action Abstractions From Language}.
\newblock In {\em {Proceedings of the International Conference on Learning Representations (ICLR)}}, 2024.

\bibitem{curtis2022discovering}
Aidan Curtis, Tom Silver, Joshua~B Tenenbaum, Tom{\'a}s Lozano-P{\'e}rez, and Leslie Kaelbling.
\newblock {Discovering State And Action Abstractions For Generalized Task And Motion Planning}.
\newblock In {\em Proceedings of The AAAI Conference On Artificial Intelligence (AAAI)}, number~5, pages 5377--5384, 2022.

\bibitem{yang2024guidinglonghorizontaskmotion}
Zhutian Yang, Caelan Garrett, Dieter Fox, Tomás Lozano-Pérez, and Leslie~Pack Kaelbling.
\newblock {Guiding Long-Horizon Task and Motion Planning with Vision Language Models}, 2024.

\bibitem{shah2024reals}
Naman Shah, Jayesh Nagpal, Pulkit Verma, and Siddharth Srivastava.
\newblock {From Reals to Logic and Back: Inventing Symbolic Vocabularies, Actions and Models for Planning from Raw Data}.
\newblock {\em arXiv preprint arXiv:2402.11871}, 2024.

\bibitem{silver2021operator}
Tom Silver, Rohan Chitnis, Joshua Tenenbaum, Leslie~Pack Kaelbling, and Tom{\'a}s Lozano-P{\'e}rez.
\newblock {Learning Symbolic Operators for Task and Motion Planning}.
\newblock In {\em {Proceedings of the IEEE/RSJ International Conference on Intelligent Robots and Systems (IROS)}}, pages 3182--3189, 2021.

\bibitem{silver2022skills}
Tom Silver, Ashay Athalye, Joshua~B. Tenenbaum, Tom{\'a}s Lozano-P{\'e}rez, and Leslie~Pack Kaelbling.
\newblock {Learning Neuro-Symbolic Skills for Bilevel Planning}.
\newblock In {\em {Proceedings of the Conference on Robot Learning (CoRL)}}, 2022.

\bibitem{silver2023predicateinvent}
Tom Silver, Rohan Chitnis, Nishanth Kumar, Willie McClinton, Tom{\'a}s Lozano-P{\'e}rez, Leslie Kaelbling, and Joshua~B Tenenbaum.
\newblock {Predicate Invention for Bilevel Planning}.
\newblock In {\em {Proceedings of The AAAI Conference on Artificial Intelligence (AAAI)}}, volume~37, pages 12120--12129, 2023.

\bibitem{chitnis2021glib}
Rohan Chitnis, Tom Silver, Joshua~B Tenenbaum, Leslie~Pack Kaelbling, and Tom{\'a}s Lozano-P{\'e}rez.
\newblock {GLIB: Efficient Exploration for Relational Model-Based Reinforcement Learning via Goal-Literal Babbling}.
\newblock In {\em {Proceedings of The AAAI Conference on Artificial Intelligence (AAAI)}}, volume~35, pages 11782--11791, 2021.

\bibitem{kumar2024practice}
Nishanth Kumar, Tom Silver, Willie McClinton, Linfeng Zhao, Stephen Proulx, Tom{\'a}s Lozano-P{\'e}rez, Leslie~Pack Kaelbling, and Jennifer Barry.
\newblock {Practice Makes Perfect: Planning To Learn Skill Parameter Policies}.
\newblock In {\em {Proceedings of the Robotics: Science And Systems (RSS)}}, 2024.

\bibitem{kumar2023predict}
Nishanth Kumar, Willie McClinton, Rohan Chitnis, Tom Silver, Tom{\'a}s Lozano-P{\'e}rez, and Leslie~Pack Kaelbling.
\newblock {Learning Efficient Abstract Planning Models That Choose What to Predict}.
\newblock In {\em {Proceedings of the Conference on Robot Learning (CoRL)}}, 2023.

\bibitem{liang2024visualpredicator}
Yichao Liang, Nishanth Kumar, Hao Tang, Adrian Weller, Joshua~B Tenenbaum, Tom Silver, João~F Henriques, and Kevin Ellis.
\newblock {VisualPredicator: Learning Abstract World Models With Neuro-Symbolic Predicates For Robot Planning}, 2024.

\bibitem{kulick2013active}
Johannes Kulick, Marc Toussaint, Tobias Lang, and Manuel Lopes.
\newblock Active learning for teaching a robot grounded relational symbols.
\newblock In {\em IJCAI}, pages 1451--1457. Citeseer, 2013.

\bibitem{konidaris2018skills}
George Konidaris, Leslie~Pack Kaelbling, and Tom{\'a}s Lozano-P{\'e}rez.
\newblock From skills to symbols: Learning symbolic representations for abstract high-level planning.
\newblock {\em Journal of Artificial Intelligence Research (JAIR)}, 2018.

\bibitem{li2023embodied}
Amber Li and Tom Silver.
\newblock {Embodied Active Learning of Relational State Abstractions for Bilevel Planning}.
\newblock In {\em {Proceedings of the Conference on Lifelong Learning Agents (CoLLAs)}}, pages 358--375, 2023.

\bibitem{han2024interpret}
Muzhi Han, Yifeng Zhu, Song-Chun Zhu, Ying~Nian Wu, and Yuke Zhu.
\newblock {InterPreT: Interactive Predicate Learning from Language Feedback for Generalizable Task Planning}.
\newblock {\em arXiv preprint arXiv:2405.19758}, 2024.

\bibitem{migimatsu2022grounding}
Toki Migimatsu and Jeannette Bohg.
\newblock {Grounding Predicates through Actions}.
\newblock In {\em Proceedings of the International Conference on Robotics and Automation (ICRA)}, pages 3498--3504. IEEE, 2022.

\bibitem{asai2018latplan_prop}
Masataro Asai and Alex Fukunaga.
\newblock {Classical Planning in Deep Latent Space: Bridging the Subsymbolic-Symbolic Boundary}.
\newblock In {\em {Proceedings of The AAAI Conference on Artificial Intelligence (AAAI)}}, volume~32, 2018.

\bibitem{asai2019latplan_fol}
Masataro Asai.
\newblock {Unsupervised Grounding of Plannable First-Order Logic Representation From Images}.
\newblock In {\em {Proceedings of the International Conference on Automated Planning and Scheduling (ICAPS)}}, volume~29, pages 583--591, 2019.

\bibitem{asai2021latplanpddl}
Masataro Asai and Christian Muise.
\newblock {Learning Neural-Dymbolic Descriptive Planning Models via Cube-Space Priors: the Voyage Home (to STRIPS)}.
\newblock In {\em Proceedings of the International Joint Conferences on Artificial Intelligence (IJCAI)}, pages 2676--2682, 2021.

\bibitem{hansen2022bisimulation}
Philippe Hansen-Estruch, Amy Zhang, Ashvin Nair, Patrick Yin, and Sergey Levine.
\newblock {Bisimulation Makes Analogies in Goal-conditioned Reinforcement Learning}.
\newblock In {\em Proceedings of the International Conference on Machine Learning (ICML)}, pages 8407--8426, 2022.

\bibitem{athalye2024predicate}
Ashay Athalye, Nishanth Kumar, Tom Silver, Yichao Liang, Tom{\'a}s Lozano-P{\'e}rez, and Leslie~Pack Kaelbling.
\newblock {Predicate Invention from Pixels via Pretrained Vision-Language Models}.
\newblock {\em arXiv preprint arXiv:2501.00296}, 2024.

\bibitem{chitnis2021nsrt}
Rohan Chitnis, Tom Silver, Joshua~B. Tenenbaum, Tom{\'a}s Lozano-P{\'e}rez, and Leslie~Pack Kaelbling.
\newblock {Learning Neuro-Symbolic Relational Transition Models for Bilevel Planning}.
\newblock In {\em {Proceedings of the IEEE/RSJ International Conference on Intelligent Robots and Systems (IROS)}}, pages 4166--4173, 2022.

\bibitem{garrett2021integrated}
Caelan~Reed Garrett, Rohan Chitnis, Rachel Holladay, Beomjoon Kim, Tom Silver, Leslie~Pack Kaelbling, and Tom{\'a}s Lozano-P{\'e}rez.
\newblock {Integrated Task and Motion Planning}.
\newblock {\em {Annual Review of Control, Robotics, and Autonomous Systems}}, 4(1):265--293, 2021.

\bibitem{helmert2006fast}
Malte Helmert.
\newblock {The Fast Downward Planning System}.
\newblock {\em {Journal of Artificial Intelligence Research}}, 26:191--246, 2006.

\bibitem{lin1991divergence}
Jianhua Lin.
\newblock {Divergence Measures based on the Shannon Entropy}.
\newblock {\em IEEE Transactions on Information theory}, 37(1):145--151, 1991.

\bibitem{kingma2014adam}
Diederik~P. Kingma and Jimmy Ba.
\newblock {Adam: A Method for Stochastic Optimization}.
\newblock {\em arXiv preprint arXiv:1412.6980}, 2014.

\bibitem{rumelhart1986sgd}
David~E Rumelhart, Geoffrey~E Hinton, and Ronald~J Williams.
\newblock {Learning Representations by Back-propagating Errors}.
\newblock {\em nature}, 323(6088):533--536, 1986.

\bibitem{coulom2006mcts}
R{\'e}mi Coulom.
\newblock {Efficient Selectivity and Backup Operators in Monte-Carlo Tree Search}.
\newblock In {\em {Proceedings of the International Conference on Computers and Games (ICCG)}}, pages 72--83, 2006.

\bibitem{silver2017alphago}
David Silver, Julian Schrittwieser, Karen Simonyan, Ioannis Antonoglou, Aja Huang, Arthur Guez, Thomas Hubert, Lucas Baker, Matthew Lai, Adrian Bolton, et~al.
\newblock {Mastering the Game of Go without Human Knowledge}.
\newblock {\em nature}, 550(7676):354--359, 2017.

\bibitem{battaglia2018gnn}
Peter~W Battaglia, Jessica~B Hamrick, Victor Bapst, Alvaro Sanchez-Gonzalez, Vinicius Zambaldi, Mateusz Malinowski, Andrea Tacchetti, David Raposo, Adam Santoro, Ryan Faulkner, et~al.
\newblock {Relational Inductive Biases, Deep Learning, and Graph Networks}.
\newblock {\em arXiv preprint arXiv:1806.01261}, 2018.

\bibitem{vaswani2017tf}
Ashish Vaswani, Noam Shazeer, Niki Parmar, Jakob Uszkoreit, Llion Jones, Aidan~N Gomez, Łukasz Kaiser, and Illia Polosukhin.
\newblock {Attention is All You Need}.
\newblock In {\em Proceedings of the Advances in Neural Information Processing Systems (NeurIPS)}, volume~30, 2017.

\bibitem{lang_sam}
Luca Medeiros.
\newblock {Language Segment-Anything}.
\newblock \url{https://github.com/luca-medeiros/lang-segment-anything}, 2024.

\bibitem{ravi2024sam}
Nikhila Ravi, Valentin Gabeur, Yuan-Ting Hu, Ronghang Hu, Chaitanya Ryali, Tengyu Ma, Haitham Khedr, Roman R{\"a}dle, Chloe Rolland, Laura Gustafson, et~al.
\newblock {Sam 2: Segment Anything in Images and Videos}.
\newblock {\em arXiv preprint arXiv:2408.00714}, 2024.

\bibitem{kaelbling2011htn}
Leslie~Pack Kaelbling and Tom{\'a}s Lozano-P{\'e}rez.
\newblock {Hierarchical Task and Motion Planning in the Now}.
\newblock In {\em Proceedings of the International Conference on Robotics and Automation (ICRA)}, pages 1470--1477. IEEE, 2011.

\bibitem{gupta2020relay}
Abhishek Gupta, Vikash Kumar, Corey Lynch, Sergey Levine, and Karol Hausman.
\newblock {Relay Policy Learning: Solving Long-Horizon Tasks via Imitation and Reinforcement Learning}.
\newblock In {\em {Proceedings of the Conference on Robot Learning (CoRL)}}, pages 1025--1037, 2020.

\bibitem{Soroush2022maple}
Soroush Nasiriany, Huihan Liu, and Yuke Zhu.
\newblock {Augmenting Reinforcement Learning with Behavior Primitives for Diverse Manipulation Tasks}.
\newblock In {\em Proceedings of the International Conference on Robotics and Automation (ICRA)}, pages 7477--7484, 2022.

\bibitem{dong2019nlm}
Honghua Dong, Jiayuan Mao, Tian Lin, Chong Wang, Lihong Li, and Denny Zhou.
\newblock {Neural Logic Machines}.
\newblock In {\em Proceedings of the International Conference on Learning Representations (ICLR)}, pages 1--10, 2019.

\bibitem{sharmadirected}
Mohit Sharma, Arjun Sharma, Nicholas Rhinehart, and Kris~M Kitani.
\newblock {Directed-Info GAIL: Learning Hierarchical Policies from Unsegmented Demonstrations using Directed Information}.
\newblock In {\em Proceedings of the International Conference on Learning Representations (ICLR)}, 2018.

\bibitem{kipf2019compile}
Thomas Kipf, Yujia Li, Hanjun Dai, Vinicius Zambaldi, Alvaro Sanchez-Gonzalez, Edward Grefenstette, Pushmeet Kohli, and Peter Battaglia.
\newblock {Compile: Compositional Imitation Learning and Execution}.
\newblock In {\em Proceedings of the International Conference on Machine Learning (ICML)}, pages 3418--3428, 2019.

\bibitem{mao2022pdsketch}
Jiayuan Mao, Tom{\'a}s Lozano-P{\'e}rez, Josh Tenenbaum, and Leslie Kaelbling.
\newblock {PDSketch: Integrated Domain Programming, Learning, and Planning}.
\newblock In {\em Proceedings of the Advances in Neural Information Processing Systems (NeurIPS)}, volume~35, pages 36972--36984, 2022.

\bibitem{liu2024BLADE}
Weiyu Liu, Neil Nie, Ruohan Zhang, Jiayuan Mao, and Jiajun Wu.
\newblock {BLADE: Learning Compositional Behaviors from Demonstration and Language}.
\newblock In {\em {Proceedings of the Conference on Robot Learning (CoRL)}}, 2024.

\bibitem{fang2024keypoint}
Xiaolin Fang, Bo-Ruei Huang, Jiayuan Mao, Jasmine Shone, Joshua~B Tenenbaum, Tom{\'a}s Lozano-P{\'e}rez, and Leslie~Pack Kaelbling.
\newblock {Keypoint Abstraction using Large Models for Object-Relative Imitation Learning}.
\newblock {\em arXiv preprint arXiv:2410.23254}, 2024.

\bibitem{silver2024generalized}
Tom Silver, Soham Dan, Kavitha Srinivas, Joshua~B Tenenbaum, Leslie Kaelbling, and Michael Katz.
\newblock {Generalized Planning in PDDL Domains with Pretrained Large Language Models}.
\newblock In {\em Proceedings of the AAAI Conference on Artificial Intelligence (AAAI)}, volume~38, pages 20256--20264, 2024.

\bibitem{wei2022cot}
Jason Wei, Xuezhi Wang, Dale Schuurmans, Maarten Bosma, Fei Xia, Ed~Chi, Quoc~V Le, Denny Zhou, et~al.
\newblock {Chain-of-Thought Prompting Elicits Reasoning in Large Language Models}.
\newblock In {\em Proceedings of the Advances in Neural Information Processing Systems (NeurIPS)}, 2022.

\bibitem{huang2023voxposer}
Wenlong Huang, Chen Wang, Ruohan Zhang, Yunzhu Li, Jiajun Wu, and Li~Fei-Fei.
\newblock {VoxPoser: Composable 3D Value Maps for Robotic Manipulation with Language Models}.
\newblock In {\em {Proceedings of the Conference on Robot Learning (CoRL)}}, pages 540--562, 2023.

\bibitem{hu2023look}
Yingdong Hu, Fanqi Lin, Tong Zhang, Li~Yi, and Yang Gao.
\newblock {Look Before You Leap: Unveiling the Power of GPT-4v in Robotic Vision-Language Planning}.
\newblock {\em arXiv preprint arXiv:2311.17842}, 2023.

\bibitem{kumar2024openworld}
Nishanth Kumar, Fabio Ramos, Dieter Fox, and Caelan~Reed Garrett.
\newblock {Open-World Task and Motion Planning via Vision-Language Model Inferred Constraints}.
\newblock In {\em CoRL Workshop on Language and Robot Learning: Language as an Interface}, 2024.

\bibitem{garrett2020pddlstream}
Caelan~Reed Garrett, Tom{\'a}s Lozano-P{\'e}rez, and Leslie~Pack Kaelbling.
\newblock {Pddlstream: Integrating Symbolic Planners and Blackbox Samplers via Optimistic Adaptive Planning}.
\newblock In {\em Proceedings of the International Conference on Automated Planning and Scheduling (ICAPS)}, volume~30, pages 440--448, 2020.

\bibitem{McDermott1998PDDL}
Drew McDermott, Malik Ghallab, Adele~E. Howe, Craig~A. Knoblock, Ashwin Ram, Manuela~M. Veloso, Daniel~S. Weld, and David~E. Wilkins.
\newblock {PDDL-the Planning Domain Definition Language}.
\newblock 1998.

\bibitem{karaman2011anytime}
Sertac Karaman, Matthew~R Walter, Alejandro Perez, Emilio Frazzoli, and Seth Teller.
\newblock {Anytime Motion Planning using the RRT}.
\newblock In {\em Proceedings of the International Conference on Robotics and Automation (ICRA)}, pages 1478--1483, 2011.

\bibitem{bougie2020skill}
Nicolas Bougie and Ryutaro Ichise.
\newblock Skill-based curiosity for intrinsically motivated reinforcement learning.
\newblock {\em Machine Learning}, 109, 2020.

\bibitem{younes2004ppddl1}
H{\aa}kan~LS Younes and Michael~L Littman.
\newblock {PPDDL1. 0: An extension to PDDL for expressing planning domains with probabilistic effects}.
\newblock {\em Techn. Rep. CMU-CS-04-162}, 2:99, 2004.

\bibitem{mao2024hybrid}
Jiayuan Mao, Joshua~B Tenenbaum, Tom{\'a}s Lozano-P{\'e}rez, and Leslie~Pack Kaelbling.
\newblock {Hybrid Declarative-Imperative Representations for Hybrid Discrete-Continuous Decision-Making}.
\newblock In {\em Proceedings of the International Workshop on the Algorithmic Foundations of Robotics (WAFR)}.

\bibitem{ravi2020pytorch3d}
Nikhila Ravi, Jeremy Reizenstein, David Novotny, Taylor Gordon, Wan-Yen Lo, Justin Johnson, and Georgia Gkioxari.
\newblock {Accelerating 3D Deep Learning with PyTorch3D}.
\newblock {\em arXiv:2007.08501}, 2020.

\bibitem{qi2017pointnet}
Charles~R Qi, Hao Su, Kaichun Mo, and Leonidas~J Guibas.
\newblock {Pointnet: Deep Learning on Point Sets for 3D Classification and Segmentation}.
\newblock In {\em Proceedings of the IEEE/CVF Conference on Computer Vision and Pattern Recognition (CVPR)}, pages 652--660, 2017.

\bibitem{he2016res}
Kaiming He, Xiangyu Zhang, Shaoqing Ren, and Jian Sun.
\newblock {Deep Residual Learning for Image Recognition}.
\newblock In {\em Proceedings of the IEEE/CVF Conference on Computer Vision and Pattern Recognition (CVPR)}, pages 770--778, 2016.

\end{thebibliography}

\newpage
\clearpage
\appendix

\subsection{Complete Notation Table}\label{app:notation}

We have presented the complete notation definition in \tref{tab:notation} for reference.

\subsection{Discussion on the Sparsity Assumption}\label{app:assumption}
Formally, we define the sparsity assumption required as:
\begin{assumption}[Effect Sparsity]
    Let $\underline{\mathtt{C}}$ be a ground action with object sets $\mathcal{O}_{\underline{\mathtt{C}}}$, $\underline{\psi}$ be a 
    ground predicate with object sets $\mathcal{O}_{\underline{\psi}}$, and $(\mathbf{x}, \underline{\mathtt{C}}, \mathbf{x}')$ be a transition pair.
    If $\mathcal{O}_{\underline{\psi}} \not\subset\mathcal{O}_{\underline{\mathtt{C}}}$, then $\theta_{\underline{\psi}}(\mathbf{x})=\theta_{\underline{\psi}}(\mathbf{x}')$.
\end{assumption}
Generally, this assumption holds if the following two conditions are both satisfied:
\begin{itemize}
    \item Each of the actions $\mathtt{C}\in\mathcal{C}$ in the domain has only one unique planning operator $\mathtt{Op}^{\mathtt{C}}$.
    \item The variable set in actions $\mathtt{C}$ is the same as the variable set in the operator $\mathtt{Op}^{\mathtt{C}}$.
\end{itemize}
In other words, if the options~\cite{chitnis2021nsrt} in a domain equal to the potential operators, then the assumption holds.
Though this is true in many domains, including those from previous work~\cite{kumar2024practice}, there exist domains where this assumption breaks~\cite{silver2023predicateinvent,chitnis2021nsrt}.
If the assumption no longer holds, more entries will become non-zero in $\bm{t}^{\psi, \underline{\mathtt{C}}}(\mathbf{x}, \mathbf{x}')$ and it would be very hard to identify their locations.
One future work that could potentially address this issue is by alternating between the predicate learning framework proposed in \sref{sec:ivntr} and the clustering mechanism proposed in previous work~\cite{chitnis2021nsrt}.

\begin{table}[!t]
    \centering
    \setlength{\tabcolsep}{1mm}
    \fontsize{8}{10}\selectfont
    \caption{List of the important notations in this work.}
    \begin{tabular}{ccc}
        \toprule[1.5pt]
        \textbf{Symbol}	& \textbf{Meaning}  & \textbf{Space}  \\
        \midrule
        $\Lambda$ & The type set in a domain & Set \\
        $\lambda$ & A type in a domain & Symbol \\
        $\mathtt{?\lambda}$ & A typed variable & Symbol \\
        $T$ & A task in a domain & Tuple \\
        $\mathcal{T}$ & Task Distribution & Distribution \\
        $K$ & Domain specific feature dimensions & Scalar \\
        $N$ & Number of objects in a task & Scalar \\
        $\mathbf{x}_i$ & The $i$-th state & Matrix \\
        $\mathcal{O}$ & Object set in a task & Set \\
        $\mathcal{C}$ & The action set in a domain & Set \\
        $\mathtt{C}$ & A parametrized action in a domain & Symbol \\
        $M$ & Number of actions in a domain & Scalar \\
        $\Omega$ & Space of the continuous action parameter & Set \\
        $\omega$ & The specific continuous action parameter & Vector \\
        $\underline{\mathtt{C}}$ & A grounded and refined action& Symbol \\
        $\underline{\bar{\mathtt{C}}}$ & A grounded but not refined action& Symbol \\
        $f$ & Transition function for a task & Function \\
        $\psi$ & Lifted predicate & Symbol \\
        $\underline{\psi}$ & Grounded predicate & Symbol \\
        $\theta_\psi$ & Classifier for predicate $\psi$ & Function \\
        $\theta_{\underline{\psi}}$ & Classification result for grounded predicate $\underline{\psi}$ & Scalar \\
        $\Psi$ & Complete predicate set for a domain & Set \\
        $\Psi_\mathrm{sta}$ & Static predicate set for a domain & Set \\
        $\Psi_\mathrm{dyn}$ & Dynamic predicate set for a domain & Set \\
        $\Psi_\mathrm{G}$ & Goal predicate set for a domain & Set \\
        $\theta_\Psi$ & Classifier set for predicate set $\Psi$ & Set \\
        $\pi$ & Refined plan & List \\
        $\bar{\pi}$ & Plan skeleton & List \\
        $\mathcal{D}$ & The dataset for learning & Set \\
        $B$ & The number of task-plan pairs & Scalar \\
        $\mathtt{Op}^\mathtt{C}$ & Operator for action $\mathtt{C}$ & Set \\
        $\mathtt{Var}$ & Variable set & List \\
        $\mathtt{Pre}$ & Pre-condition set & Set \\
        $\mathtt{Eff}^+$ & Add effect set & Set \\
        $\mathtt{Eff}^-$ & Delete effect set & Set \\
        $\eta^\mathtt{C}$ & Sampler for action $\mathtt{C}$ & Function \\
        $\hat{\Psi}_\mathrm{dyn}$ & Candidate predicate set & Set \\
        $J(\cdot)$ & Score function based on planning outcome & Function \\
        $\Psi^\mathtt{Var}$ & Set of predicates with typed variables $\mathtt{Var}$ & Set \\
        $\Delta^\psi$ & Effect vector for predicate $\psi$ & Vector \\
        $\delta^\psi_i$ & Effect value for predicate $\psi$ in action $\mathtt{C}_i$& Scalar \\
        $\bm{t}^{\psi, \underline{\mathtt{C}}}$ & Predicted ground predicate transition & Vector \\
        $\Delta^{\Psi^\mathtt{Var}}$ & Effect vector set for predicate set $\Psi^\mathtt{Var}$ & Set \\
        $\mathbf{L}_t$ & Loss information at $t$-th iteration & Vector \\
        $r^n_t$ & Value for the $n$-th effect vector at $t$-th iteration & Scalar \\
        $\mathbf{R}_t$ & Global value vector at $t$-th iteration & Vector \\
        \bottomrule[1.5pt]
    \end{tabular}%
    \label{tab:notation}%
    \vspace{-0.3cm}
\end{table}%

\subsection{Ground Predicates with Action Variables}\label{app:same_type}
In general, for a transition with ground action $\underline{\mathtt{C}}$, multiple ground predicates could be ``grounded on" the object set $\mathcal{O}_{\underline{\mathtt{C}}}$.
For example, in the Blocks domain, we could have actions like $\mathtt{Stack(?r,?b_0,?b_1)}$, where there are two same typed variables $\mathtt{?b_0,?b_1}$ (two objects in the blocks type).
In this case, consider the object set $\{\mathtt{r_1,b_1,b_2}\}$, if we have ground action $\mathtt{Stack(r_1,b_1,b_2)}$, then the two ground predicates $\mathtt{P1(b_1,b_2)}$ and $\mathtt{P1(b_2,b_1)}$ should be both considered as ``grounded on" the object set $\mathcal{O}_{\underline{\mathtt{C}}}$.
In such cases, the two ground predicates will have the same transition following \defref{def:ground_effect_vec}.
However, this is usually not true, for example, if $\mathtt{P1(?b,?b)}=\mathtt{On(?b,?b)}$, then the transition of $\mathtt{On(b_1,b_2)}$ and $\mathtt{On(b_2,b_1)}$ should \textbf{not} be the same.
To solve this problem, when we define the predicate sets $\Psi^\mathtt{Var}$ with the same typed variables $\mathtt{Var}$, we additionally annotate that the predicate variables $\mathtt{Var}$ have a fixed correspondence with the action variables $\mathtt{Var}_\mathtt{c}$.
More specifically, when we ground the predicates, all possible object compositions are considered.
But when we use the ground predicates to form effects (during predicate function training) and pre-conditions (during operator learning), we only use the ground predicates whose object sets have the specified correspondence with the action variables.
For example, consider the predicate $\mathtt{P1(?b,?b)}$, we may further annotate it with a list $[0,1]$, meaning that the first predicate variables should be matched with the first variable in the actions that is typed as a block and the second predicate variables should be matched with the second block variable in the actions.
In this case, we will only consider $\mathtt{P1(b_1,b_2)}$ for transitions with action $\mathtt{Stack(r_1,b_1,b_2)}$ (and $\mathtt{P1(b_2,b_1)}$ with action $\mathtt{Stack(r_1,b_2,b_1)}$).
In practice, the predicate sets with the same typed variables but different correspondence annotation are considered as different groups in the bilevel learning.
For more details, please refer to our source code.

\subsection{Improving Efficiency during Expansion}\label{app:pruning}
To make the tree expansion (symbolic learning) more efficient without sacrificing too much completeness, we have tried to prune an effect vector $\Delta^{\psi_n}$ (setting its value $r^n$ as $-\infty$) via the following strategy:
\begin{itemize}
    \item If the input object-centric states to the predicate function $\theta_{\psi_n}$ never change for any ground actions $\underline{\mathtt{C}}$ belonging to an action $\mathtt{C}$, then the nodes $\Delta^{\psi_n}$ with non-zero entry $\Delta^{\psi_n}_\mathtt{C}$ will be pruned. This strategy is implemented as a ``pre-check", which happens before the tree expansion starts.
    \item Assume the loss vector $\mathbf{L}_t$ is from the evaluated effector vector $\Delta^\psi_t$ in the $t$-th iteration. Let $\mathbb{I}(\Delta^{\psi}_t \neq 0)$ and $\mathbb{I}(\Delta^{\psi_n} \neq 0)$ be the mask indicating whether the entries in $\Delta^\psi_t$ and $\Delta^{\psi_n}$ are non-zero, respectively. We prune $\Delta^{\psi_n}$ if $\sum \mathbf{L}_t\odot\mathbb{I}(\Delta^{\psi}_t \neq 0)\odot\mathbb{I}(\Delta^{\psi_n} \neq 0) > \tau$.
\end{itemize}
The second strategy prunes a vector if the sum of the loss from its non-zero indices is larger than the threshold. 
This strategy might make the tree expansion not fully complete. 
The intuition behind it is that: Since the children have the same non-zero parts as their parents, if the parents' non-zero part has already contributed high validation loss, then we assume the children's non-zero part will also contribute high loss. 
There is a small chance that the children turn out to have low loss due to the additional non-zero entry.
We have empirically shown that this strategy works in practice, probably due to the fact that a child node can come from different parents.

\begin{figure*}[!t]
	\centering
	\includegraphics[width=2\columnwidth]{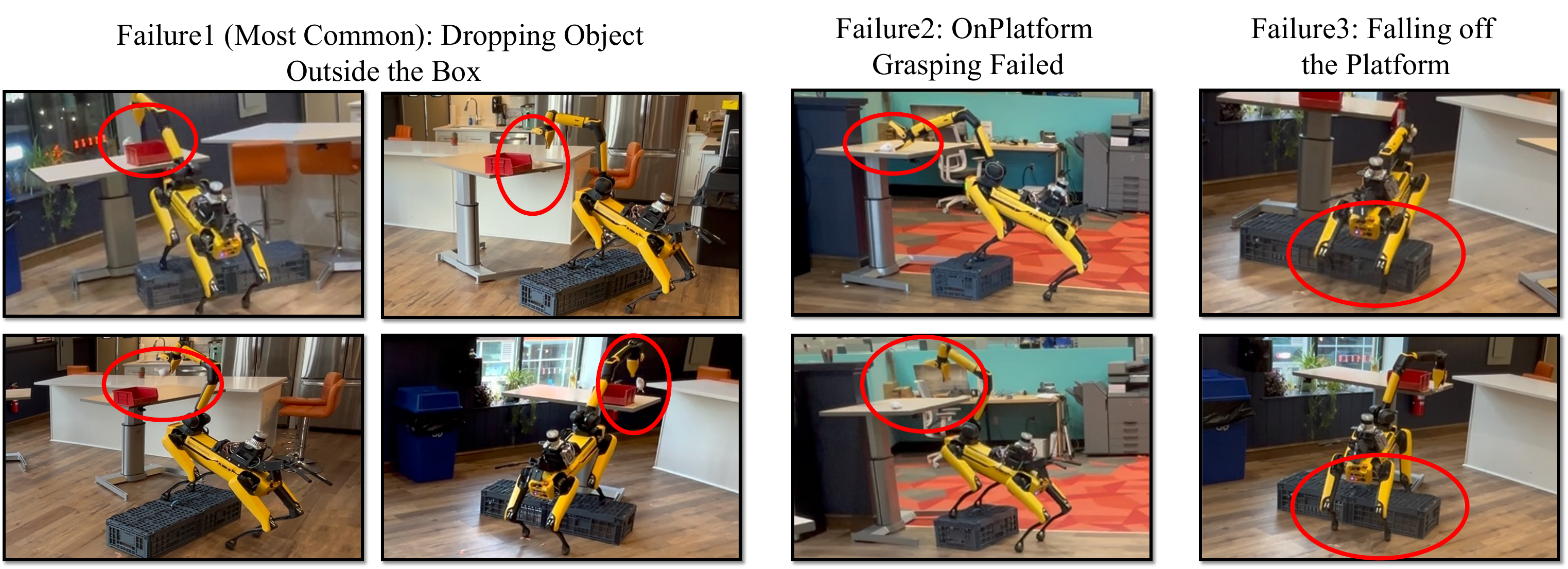}
 \vspace{-0.1cm}
	\caption{Typical failure cases in the real robot tests of the Climb-Transport domain. The most common failure is that the target is dropped outside of the container, which is due to the localization error. Other failures include failing to generate the grasping motion plan on the platform and falling of the platform while trying to climb onto it.}
 \vspace{-0.1cm}
	\label{fig:app_failure}
\end{figure*}

\subsection{Implementation Details for Each Domain}\label{app:domain_details}

Here, we provide more details for each domain:

\textbf{Satellites:}
\begin{itemize}
    \item \textit{Types}: The satellites ($\mathtt{s}$) and the targets ($\mathtt{t}$). 
    \item \textit{Actions}: $\mathtt{Calibrate(?s,?t)}$, $\mathtt{MoveTo(?s,?t)}$, $\mathtt{MoveAway(?s,?t)}$, $\mathtt{ShootX(?s,?t)}$, $\mathtt{ShootY(?s,?t)}$, $\mathtt{TakeCam(?s,?t)}$, $\mathtt{TakeInfrared(?s,?t)}$, and $\mathtt{TakeGeiger(?s,?t)}$.
    \item \textit{Static Predicates}: $\mathtt{CalibrationTgt(?s,?t)}$, $\mathtt{ShootsX(?s)}$, $\mathtt{ShootsY(?s)}$.
    \item \textit{Goal Predicates}: $\mathtt{CamTaken(?s,?t)}$, $\mathtt{InfraredTaken(?s,?t)}$, $\mathtt{GeigerTaken(?s,?t)}$.
    \item \textit{Task Description}: There are some number of satellites, each carrying an instrument. The possible instruments are: (1) a camera, (2) an infrared sensor, (3) a Geiger counter.
    Additionally, each satellite may be able to shoot Chemical X and/or Chemical
    Y. The satellites have a viewing cone within which they can see everything
    that is not occluded. The goal is for specific satellites to take readings
    of specific objects with calibrated instruments.
    \item \textit{Predicate invention hardware}: A single A100 GPU.
\end{itemize}

\textbf{Blocks:}
\begin{itemize}
    \item \textit{Types}: The robot ($\mathtt{r}$) and the blocks ($\mathtt{b}$). 
    \item \textit{Actions}: $\mathtt{PickFromTable(?r,?b)}$, $\mathtt{Unstack(?r,?b,?b)}$, $\mathtt{Stack(?r,?b,?b)}$, $\mathtt{PutOnTable(?r,?b)}$, $\mathtt{Pack(?b,?b)}$.
    \item \textit{Static Predicates}: None.
    \item \textit{Goal Predicates}: $\mathtt{Packed(?b,?b)}$.
    \item \textit{Task Description}: The robot needs to manipulate a set of blocks (which were initialized as random towers) and pack them into required pairs. In order to pack two blocks, one block needs to be on the top of another with the bottom block on the table and the top block clear.
    \item \textit{Predicate invention hardware}: A single A100 GPU.
\end{itemize}

\textbf{Measure-Mul:}
\begin{itemize}
    \item \textit{Types}: The robot ($\mathtt{r}$) and the targets ($\mathtt{t}$). 
    \item \textit{Actions}: $\mathtt{Calibrate(?r,?t)}$, $\mathtt{MoveTo(?r,?t)}$, $\mathtt{MoveAway(?r,?t)}$, $\mathtt{Gaze(?r,?t)}$, and $\mathtt{Measure(?r,?t)}$.
    \item \textit{Static Predicates}: $\mathtt{CalibrationTgt(?s,?t)}$.
    \item \textit{Goal Predicates}: $\mathtt{Measured(?t)}$.
    \item \textit{Task Description}: The Spot robot needs to use a thermal camera under its hand to measure the body temperature of multiple human targets.
    To do this, it needs to first calibrate the camera with respect to a calibrator by gazing at it, which poses some constraints on the relative poses between the hand and the target.
    Then, before measuring each target, it also needs to gaze at them.
    \item \textit{Predicate invention hardware}: A single A100 GPU.
\end{itemize}

\textbf{Climb-Measure:}
\begin{itemize}
    \item \textit{Types}: The robot ($\mathtt{r}$), the targets ($\mathtt{t}$), and the platform ($\mathtt{p}$).
    \item \textit{Actions}: $\mathtt{Calibrate(?r,?t)}$, $\mathtt{MoveToGaze(?r,?t)}$, $\mathtt{MoveToReach(?r,?p)}$, $\mathtt{MoveToPlace(?r,?p,?t)}$, $\mathtt{MoveAwayOff(?r,?t)}$, $\mathtt{MoveAwayOn(?r,?p,?t)}$, $\mathtt{WalkOn(?r,?p,?t)}$, $\mathtt{Pick(?r,?p)}$, $\mathtt{Place(?r,?p,?t)}$, $\mathtt{Gaze(?r,?t)}$, and $\mathtt{Measure(?r,?t)}$.
    \item \textit{Static Predicates}: $\mathtt{CalibrationTgt(?s,?t)}$, $\mathtt{DirectViewable(?t)}$, $\mathtt{AppliedTo(?p,?t)}$.
    \item \textit{Goal Predicates}: $\mathtt{Measured(?t)}$.
    \item \textit{Task Description}: Similar to Measure-Mul, the Spot robot needs to use a thermal camera under its hand to measure the body temperature of a human target.
    Differently, the calibrator and human target could be at a high location where the Spot can directly gaze at them.
    To achieve the goal, the Spot will need to arrange some platforms.
    \item \textit{Predicate invention hardware}: Parallelly on four A100 GPUs.
\end{itemize}

\textbf{Climb-Transport:}
\begin{itemize}
    \item \textit{Types}: The robot ($\mathtt{r}$), the targets ($\mathtt{t}$), and the platform ($\mathtt{p}$).
    \item \textit{Actions}: $\mathtt{Grasp(?r,?t)}$, $\mathtt{MoveToGaze(?r,?t)}$, $\mathtt{MoveToReach(?r,?p)}$, $\mathtt{MoveToPlace(?r,?p,?t)}$, $\mathtt{MoveAwayOff(?r,?p)}$, $\mathtt{MoveAwayOn(?r,?p,?t)}$, $\mathtt{WalkOn(?r,?p,?t)}$, $\mathtt{Pick(?r,?p)}$, $\mathtt{Gaze(?r,?t)}$, and $\mathtt{Drop(?r,?t,?t)}$.
    \item \textit{Static Predicates}: $\mathtt{GraspingTgt(?t)}$, $\mathtt{InitHigh(?t)}$.
    \item \textit{Goal Predicates}: $\mathtt{In(?t,?t)}$.
    \item \textit{Task Description}: The Spot robot needs to grasp a target and drop it into another target container.
    Similar to Climb-Measure, to achieve the goal, the Spot will need to arrange some platforms.
    \item \textit{Predicate invention hardware}: Parallelly on four A100 GPUs.
\end{itemize}

\textbf{Engrave:}
\begin{itemize}
    \item \textit{Types}: The robot ($\mathtt{r}$) and the blocks ($\mathtt{b}$). 
    \item \textit{Actions}: $\mathtt{PickFromTable(?r,?b)}$, $\mathtt{Unstack(?r,?b,?b)}$, $\mathtt{Stack(?r,?b,?b)}$, $\mathtt{PutOnTable(?r,?b)}$, $\mathtt{Engrave(?r, ?b,?b)}$, $\mathtt{Rotate(?r,?b)}$, $\mathtt{Pack(?b,?b)}$.
    \item \textit{Static Predicates}: $\mathtt{NotEq(?b,?b)}$
    \item \textit{Goal Predicates}: $\mathtt{Packed(?b,?b)}$.
    \item \textit{Task Description}: Similar to Blocks, the robot needs to manipulate a set of blocks (which were initialized as random towers) and pack them into required pairs. 
    However, blocks start with one irregular Gaussian surface that must be ``engraved" to create a matching fit. 
    Once engraved, blocks need to be further rotated and placed for final assembly and packing. 
    We generate the Block meshes and point clouds using Pytorch3D~\cite{ravi2020pytorch3d}. 
    For fairness, all methods use the same PointNet~\cite{qi2017pointnet} as the state encoder. 
    \item \textit{Predicate invention hardware}: Parallelly on six A100 GPUs.
\end{itemize}

For more details about these domains, please refer to our source code.

\subsection{More Details about Baselines}\label{app:baseline_details}
We introduce more details for the relational neural policy baselines here:
\begin{itemize}
    \item \textbf{GNN}: The nodes are defined the object-centric features and the edges are defined as the grounded binary predicates (provided static and goal predicates).
    During training, the GNN learns to predict the action class and the selected objects.
    During test, the GNN tries to shoot multiple tries until planning budget run out.
    \item \textbf{Transformer}: The tokens are defined the object-centric features as well as the grounded predicates (provided static and goal predicates).
    The training objective and test setup are similar to GNN.
    However, different from the message passing strategy in GNN, we used the multi-head attention mechanism for the information fusing among tokens
    \item \textbf{FOSAE}: The baseline is loosely inspired by the state autoencoder (SAE) proposed in LatPlan~\cite{asai2019latplan_fol}. We first use the official attention mechanism to pre-train the SAE module by reconstructing the original state.
    Then, since the action spaces in this work is relational (instead of nullary), we encode the augmented binary states from SAE as the global feature of a graph neural network.
    Finally, the graph neural network is trained and evaluated similar to the GNN-shooting baseline.
\end{itemize}
For fairness, we have used the same action samplers learned using our framework for the three baseline above. 
For more details, please refer to our source code.

\subsection{Failures in Real Robot Experiments}\label{app:failures}

As shown in \fref{fig:app_failure}, we present the typical failure cases in the real-robot tests for the Climb-Transport domain.
The most frequent failure is that the robot finally drops the target outside of the container (but very close).
The reason is that the map of our experiments was recorded before the tables (used to place targets) were placed, which could result in the small drift of the localization system on Spot during the plan execution.
These drifts finally accumulate and result in the final hand pose error.
A worse problem is that due to the lack of a motion capture system, the container itself might not be accurately placed.
One potential way to address this is by integrating some more advanced SLAM system that is robust to partial map changes.
Another failure case (much less common) is that the Spot grasping skill could fail when it is on a platform.
Here, we have used the manipulation toolkit from the official Boston Dynamics Python SDK, which might fail to find a grasping motion plan in certain states.
Finally, since the platform is narrow, we have also observed that the Spot could fall off the platform when walking onto the platform.

\subsection{Invented Predicates and Operators}\label{app:complete_op}
We present the invented predicates together with the operators for the Satellites domain below.
The invented predicates are named as $\mathtt{P1},\mathtt{P2},\cdots$, while the provided predicates have the names introduced in \appref{app:domain_details}.
For other domains, the learned operators are more sophisticated; please refer to our source code and meta results for details.

\textbf{Satellites:}
\begin{Verbatim}[frame=single,resetmargins=true]
Calibrate(?x0:s, ?x1:t)
  :Pre (and 
      (CalibrationTgt(?x0, ?x1))
      (not P3_0(?x0, ?x1))) 
  :Eff+ (P1_0(?x0))
  :Eff- set()
MoveAway(?x0:s, ?x1:t)
  :Pre (not P3_0(?x0, ?x1))
  :Eff+ (ForAll:?t P3_0(?x0, ?t))
  :Eff- (not P3_0(?x0, ?x1))
MoveTo(?x0:s, ?x1:t)
  :Pre (ForAll:?t P3_0(?x0, ?t))
  :Eff+ (not P3_0(?x0, ?x1))
  :Eff- (ForAll:?t P3_0(?x0, ?t))
ShootChemX(?x0:s, ?x1:t)
  :Pre (and
    (not P3_0(?x0, ?x1))
    ShootsChemX(?x0))
  :Eff+ (P2_0(?x1))
  :Eff- set()
ShootChemY(?x0:s, ?x1:t)
  :Pre (and
    (not P3_0(?x0, ?x1))
    ShootsChemY(?x0))
  :Eff+ (P2_1(?x1))
  :Eff- set()
UseCamera(?x0:s, ?x1:t)
  :Pre (and
    (not P3_0(?x0, ?x1))
    (P1_0(?x0))
    (P2_0(?x1))
    HasCam(?x0))
  :Eff+ (CameraReadingTaken(?x0, ?x1))
  :Eff- set()
UseGeiger(?x0:s, ?x1:t)
  :Pre (and
    (not P3_0(?x0, ?x1))
    (P1_0(?x0))
    HasGeiger(?x0))
  :Eff+ (GeigerReadingTaken(?x0, ?x1))
  :Eff- set()
UseInfraRed(?x0:s, ?x1:t)
  :Pre (and
    (not P3_0(?x0, ?x1))
    (P1_0(?x0))
    (P2_1(?x1))
    HasInfrared(?x0))
  :Eff+ (InfraredReadingTaken(?x0, ?x1))
  :Eff- set()\end{Verbatim}

\begin{figure}[!t]
	\centering
	\includegraphics[width=1\columnwidth]{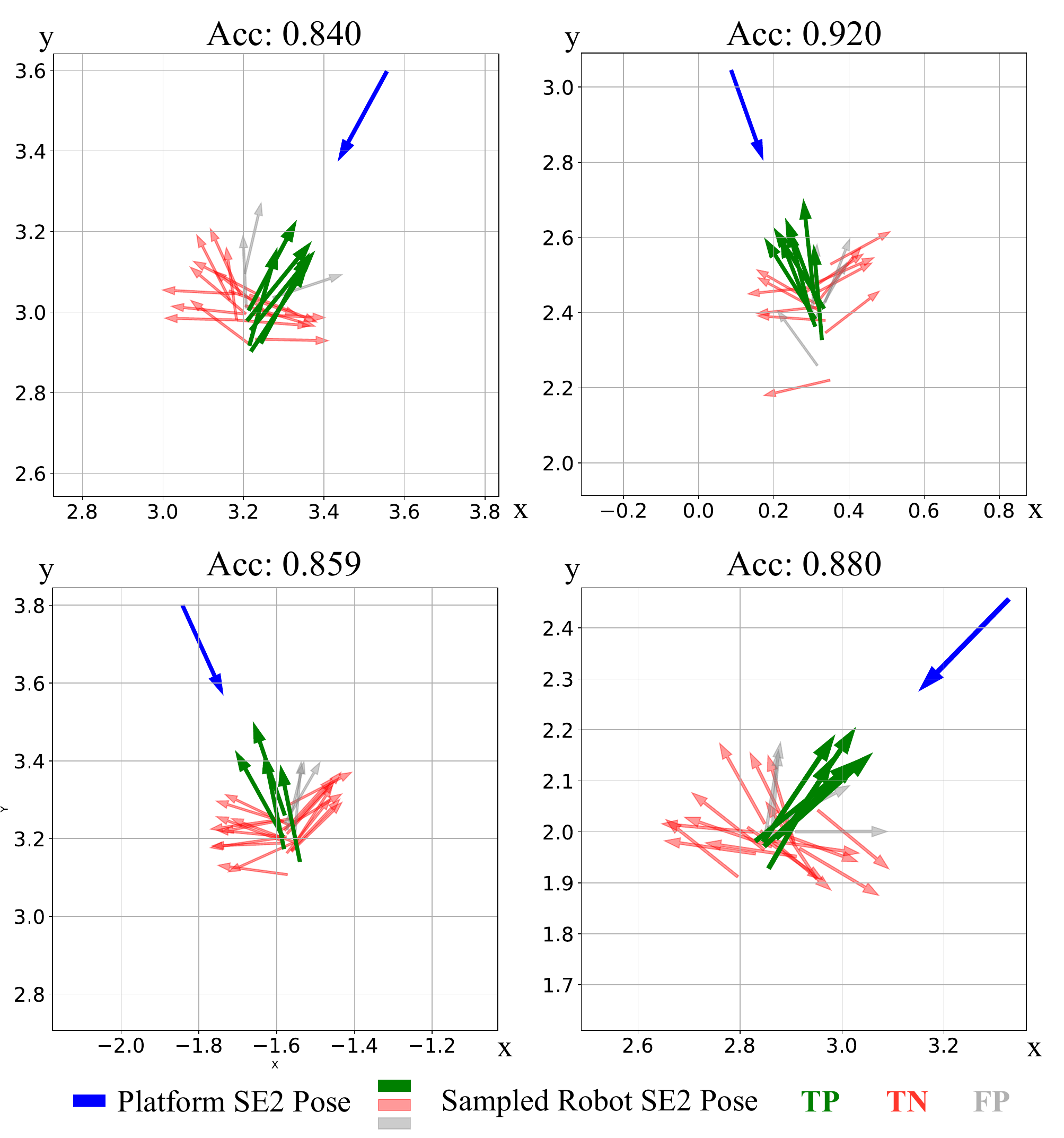}
	\caption{Visualization of the classification results on sampled poses for \texttt{MoveToReach} action using our invented predicate $\mathtt{P3(?r,?p)}$. TP is for true positive, TN is for true negatives, and FP is for false positive. We only visualize part of the samples which are converted to SE2 for simplicity, best viewed in color.}
	\label{fig:vis_sample}
\end{figure}

\subsection{Neural Predicates as Guidance for the Samplers}\label{app:sampler_vis} 
As shown in \fref{fig:vis_sample}, we visualize classification results for sampled continuous parameters of the \texttt{MoveToReach} action across $4$ tasks. 
Since training demonstrations involve solving inverse kinematics (IK) for \texttt{MoveToReach}, accurate sampler learning is challenging.
Our invented predicate $\mathtt{P3(?r,?p)}$, an add effect of \texttt{MoveToReach} and a precondition for \texttt{Pick}, effectively filters out most true negatives (TNs), ensuring successful plans.

\begin{figure}[!t]
	\centering
	\includegraphics[width=1\columnwidth]{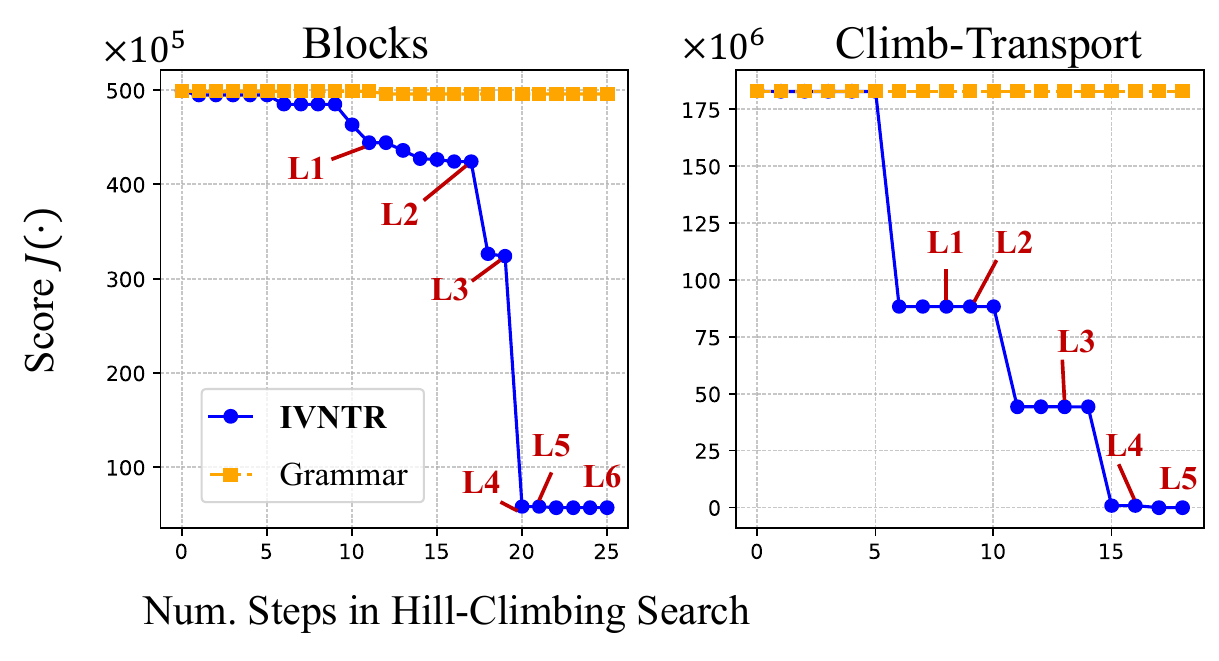}
 \vspace{-0.4cm}
	\caption{Score optimization process during the predicate selection stage. Compared with grammar based predicate pool~\cite{silver2023predicateinvent}, our \model{} is capable of constructing a much stronger neural predicate pool, which is able to effectively optimize the planning objective with a few hill-climbing steps. Yet, the grammar-based approach has failed.
    }
    \vspace{-0.4cm}
	\label{fig:ana_objective}
\end{figure}

\subsection{Objective Minimization}\label{app:objective}
We display the objective minimization process during the predicate selection stage in \fref{fig:ana_objective}.
Compared with the grammar-based predicate pool~\cite{silver2023predicateinvent}, \model{} is capable of generating much stronger neural functions as predicate candidates.
These neural predicates have made the objective minimization possible for more complicated states, where previous approaches have failed~\cite{silver2023predicateinvent}.

\subsection{Operator and Sampler Learning}\label{app:op_sam_learning}
We primarily follow existing work for sampler and operator learning~\cite{silver2023predicateinvent}.
For each predicate in the final set $\psi\in\Psi$, we obtain its lifted effect vector $\Delta^\psi$ by running the classifier on training dataset.
For the $m-$th operator $\mathtt{Op}^\mathtt{C}$, its effects are computed by checking the $m-$th entry of the effect vectors from all predicates.
The preconditions for each operator are determined by finding the intersection among all ground atoms that are true in previous states of ground transitions, known as intersection~\cite{silver2023predicateinvent}.
For sampler learning, we leverage supervised learning to train a Gaussian regressor as generator and an MLP as classifier.

\subsection{Applying IVNTR to Domains with Image States}\label{app:blocks_img}
\begin{figure}[!t]
	\centering
	\includegraphics[width=1\columnwidth]{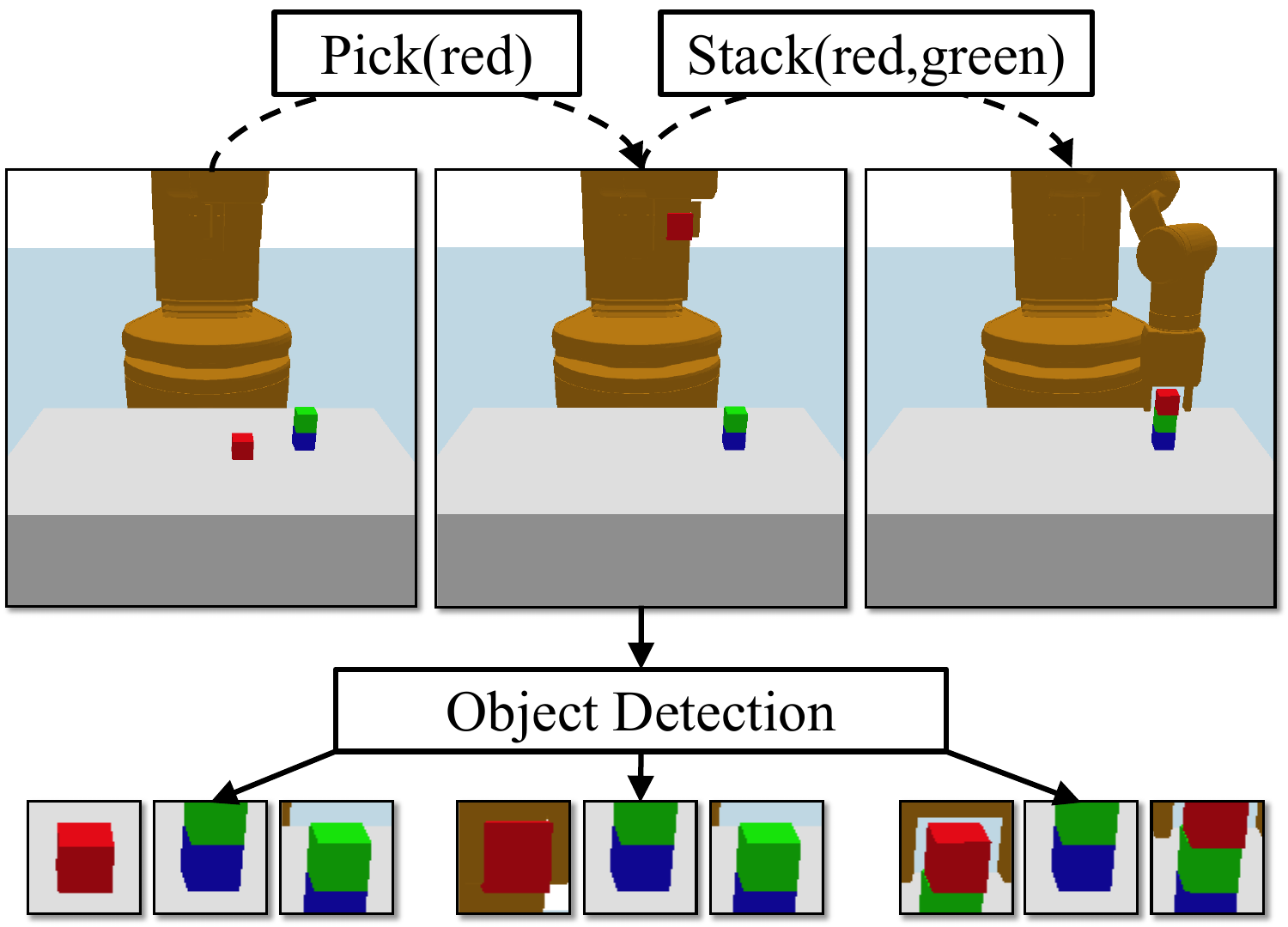}
	\caption{An example demonstration with $3$ blocks, where the states are represented in RGB images. In each step, we have used an object detection algorithm to obtain object centric images as the input states to our IVNTR algorithm.
    }
	\label{fig:blocks_img}
\end{figure}

\begin{table}[!t]
\centering
\setlength{\tabcolsep}{2.4mm}
    \fontsize{8}{10}\selectfont
    \begin{tabular}{ccccccc}
        \toprule[1.5pt]
        Method/Predicate & P1 & P2 & P3 & P4 & P5 & Avg. \\
        \midrule
        \textbf{IVNTR (Ours)} & 6 & 55 & 6 & 51 & 2 & 24 \\
        Greedy & 77 & 34 & 139 & 20 & 12 & 56.4 \\
        BFS & 15 & 30 & 55 & 218 & 14 & 66.4 \\
        DFS & 4 & 39 & 188 & 91 & 18 & 68 \\
        Random & 80 & 80 & 242 & 242 & 26 & 134 \\ 
        \bottomrule[1.5pt]
    \end{tabular}%
    \caption{Comparison between our neural guided tree expansion and other alternative tree expansion algorithms in the Climb-Transport domain. Our IVNTR has demonstrated the best average efficiency in finding the five predicates.}
  \label{tab:app_search}%
\end{table}%
We have further implemented a Blocks stacking domain with RGB images as states, see~\fref{fig:blocks_img} for visualizations.
Methods are required to generalize from $3-4$ blocks to $5$ blocks.
IVNTR was built upon ResNet18~\cite{he2016res}, which has successfully invented visual predicates like \texttt{Holding(?b)}.
The average success rate on generalized tasks of \emph{IVNTR}/GNN/No\_invention are \emph{92.0\%}/14.7\%/0.0\%.

\subsection{Search Efficiency Comparison}\label{app:more_search}
We further present the search ablations on the Climb-Transport domain in \tref{tab:app_search}, where the average search iterations using \emph{IVNTR}/Greedy/BFS/DFS/Random are \emph{24}/56.4/66.4/68/134. 

\end{document}